# Accelerating Reinforcement Learning through Implicit Imitation

**Bob Price**                                             PRICE@CS.UBC.CA
*Department of Computer Science*
*University of British Columbia*
*Vancouver, B.C., Canada V6T 1Z4*

**Craig Boutilier**                                       CEBLY@CS.TORONTO.EDU
*Department of Computer Science*
*University of Toronto*
*Toronto, ON, Canada M5S 3H5*

## Abstract

Imitation can be viewed as a means of enhancing learning in multiagent environments. It augments an agent's ability to learn useful behaviors by making intelligent use of the knowledge implicit in behaviors demonstrated by cooperative teachers or other more experienced agents. We propose and study a formal model of *implicit imitation* that can accelerate reinforcement learning dramatically in certain cases. Roughly, by observing a mentor, a reinforcement-learning agent can extract information about its own capabilities in, and the relative value of, unvisited parts of the state space. We study two specific instantiations of this model, one in which the learning agent and the mentor have identical abilities, and one designed to deal with agents and mentors with different action sets. We illustrate the benefits of implicit imitation by integrating it with prioritized sweeping, and demonstrating improved performance and convergence through observation of single and multiple mentors. Though we make some stringent assumptions regarding observability and possible interactions, we briefly comment on extensions of the model that relax these restrictions.

## 1. Introduction

The application of reinforcement learning to multiagent systems offers unique opportunities and challenges. When agents are viewed as independently trying to achieve their own ends, interesting issues in the interaction of agent policies (Littman, 1994) must be resolved (e.g., by appeal to equilibrium concepts). However, the fact that agents may share information for mutual gain (Tan, 1993) or distribute their search for optimal policies and communicate reinforcement signals to one another (Mataric, 1998) offers intriguing possibilities for accelerating reinforcement learning and enhancing agent performance.

Another way in which individual agent performance can be improved is by having a novice agent learn reasonable behavior from an expert *mentor*. This type of learning can be brought about through explicit *teaching* or demonstration (Atkeson & Schaal, 1997; Lin, 1992; Whitehead, 1991a), by sharing of privileged information (Mataric, 1998), or through an explicit cognitive representation of *imitation* (Bakker & Kuniyoshi, 1996). In imitation, the agent's own exploration is used to ground its observations of other agents' behaviors in its own capabilities and resolve any ambiguities in observations arising from partial observability and noise. A common thread in all of this work is the use of a mentor to guide the *exploration* of the observer. Typically, guidance is achieved through some form of explicit communication between mentor and observer. A less direct form of teaching involves an observer extracting information from a mentor without the mentor making an explicit attempt to demonstrate a specific behavior of interest (Mitchell, Mahadevan, & Steinberg, 1985).

In this paper we develop an imitation model we call *implicit imitation* that allows an agent to accelerate the reinforcement learning process through the observation of an expert mentor (or mentors). The agent observes the state transitions induced by the mentor's actions and uses the information gleaned from these observations to update the estimated value of its own states and actions. We will distinguish two settings in which implicit imitation can occur: *homogeneous settings*, in which the learning agent and the mentor have identical actions; and *heterogeneous* settings, where their capabilities may differ. In the homogeneous setting, the learner can use the observed mentor transitions directly to update its own estimated model of its actions, or to update its value function. In addition, a mentor can provide hints to the observer about the parts of the state space on which it may be worth focusing attention. The observer's attention to an area might take the form of additional exploration of the area or additional computation brought to bear on the agent's prior beliefs about the area. In the heterogeneous setting, similar benefits accrue, but with the potential for an agent to be misled by a mentor that possesses abilities different from its own. In this case, the learner needs some mechanism to detect such situations and to make efforts to temper the influence of these observations.

We derive several new techniques to support implicit imitation that are largely independent of any specific reinforcement learning algorithm, though they are best suited for use with model-based methods. These include model extraction, augmented backups, feasibility testing, and $k$-step repair. We first describe implicit imitation in homogeneous domains, then we describe the extension to heterogeneous settings. We illustrate its effectiveness empirically by incorporating it into Moore and Atkeson's (1993) prioritized sweeping algorithm.

The implicit imitation model has several advantages over more direct forms of imitation and teaching. It does not require any agent to explicitly play the role of mentor or teacher. Observers learn simply by watching the behavior of other agents; if an observed "mentor" shares certain subtasks with the observer, the observed behavior can be incorporated (indirectly) by the observer to improve its estimate of its own value function. This is important because there are many situations in which an observer can learn from a mentor that is unwilling or unable to alter its behavior to accommodate the observer, or even communicate information to it. For example, common communication protocols may be unavailable to agents designed by different developers (e.g., Internet agents); agents may find themselves in a competitive situation in which there is disincentive to share information or skills; or there may simply be no incentive for one agent to provide information to another.[1]

Another key advantage of our approach—which arises from formalizing imitation in the reinforcement learning context—is the fact that the observer is not constrained to directly imitate (i.e., duplicate the actions of) the mentor. The learner can decide whether such "explicit imitation" is worthwhile. Implicit imitation can thus be seen as blending the advantages of explicit teaching or explicit knowledge transfer with those of independent learning. In addition, because an agent learns by observation, it can exploit the existence of multiple mentors, essentially distributing its search. Finally, we do not assume that the observer knows the actual actions taken by the mentor, or that the mentor shares a reward function (or goals) with the mentor. Again, this stands in sharp contrast with many existing models of teaching, imitation, and behavior learning by observation. While we make some strict assumptions in this paper with respect to observability, complete knowledge of reward functions, and the existence of mappings between agent state spaces, the model can be generalized in interesting ways. We will elaborate on some of these generalizations near the end of the paper.

---

1. For reasons of consistency, we will use the term "mentor" to describe any agent from which an observer can learn, even if the mentor is an unwilling or unwitting participant.

The remainder of the paper is structured as follows. We provide the necessary background on Markov decision processes and reinforcement learning for the development of our implicit imitation model in Section 2. In Section 3, we describe a general formal framework for the study of implicit imitation in reinforcement learning. Two specific instantiations of this framework are then developed. In Section 4, a model for homogeneous agents is developed. The *model extraction* technique is explained and the *augmented Bellman backup* is proposed as a mechanism for incorporating observations into model-based reinforcement learning algorithms. Model confidence testing is then introduced to ensure that misleading information does not have undue influence on a learner's exploration policy. The use of mentor observations to to focus attention on interesting parts of the state space is also introduced. Section 5 develops a model for heterogeneous agents. The model extends the homogeneous model through *feasibility testing*, a device by which a learner can detect whether the mentor's abilities are similar to its own, and *k-step repair*, whereby a learner can attempt to "mimic" the trajectory of a mentor that cannot be duplicated exactly. Both of these techniques prove crucial in heterogeneous settings. The effectiveness of these models is demonstrated on a number of carefully chosen navigation problems. Section 6 examines conditions under which implicit imitation will and will not work well. Section 7 describes several promising extensions to the model. Section 8 examines the implicit imitation model in the context of related work and Section 9 considers future work before drawing some general conclusions about implicit imitation and the field of computational imitation more broadly.

## 2. Reinforcement Learning

Our aim is to provide a formal model of implicit imitation, whereby an agent can learn how to act optimally by combining its own experience with its observations of the behavior of an expert mentor. Before doing so, we describe in this section the standard model of reinforcement learning used in artificial intelligence. Our model will build on this single-agent view of learning how to act. We begin by reviewing Markov decision processes, which provide a model for sequential decision making under uncertainty, and then move on to describe reinforcement learning, with an emphasis on model-based methods.

## 2.1 Markov Decision Processes

Markov decision processes (MDPs) have proven very useful in modeling stochastic sequential decision problems, and have been widely used in decision-theoretic planning to model domains in which an agent's actions have uncertain effects, an agent's knowledge of the environment is uncertain, and the agent can have multiple, possibly conflicting objectives. In this section, we describe the basic MDP model and consider one classical solution procedure. We do not consider action costs in our formulation of MDPs, though these pose no special complications. Finally, we make the assumption of *full observability*. *Partially observable MDPs* (POMDPs) (Cassandra, Kaelbling, & Littman, 1994; Lovejoy, 1991; Smallwood & Sondik, 1973) are much more computationally demanding than fully observable MDPs. Our imitation model will be based on a fully observable model, though some of the generalizations of our model mentioned in the concluding section build on POMDPs. We refer the reader to Bertsekas (1987); Boutilier, Dean and Hanks (1999); and Puterman (1994) for further material on MDPs.

An MDP can be viewed as a stochastic automaton in which actions induce transitions between states, and rewards are obtained depending on the states visited by an agent. Formally, an MDP can be defined as a tuple $\langle \mathcal{S}, \mathcal{A}, T, R \rangle$, where $\mathcal{S}$ is a finite set of *states* or possible worlds, $\mathcal{A}$ is a finite set of *actions*, $T$ is a *state transition function*, and $R$ is a *reward function*. The agent can control the state of the system to some extent by performing actions $a \in \mathcal{A}$ that cause *state transitions*, movement from the current state to some new state. Actions are stochastic in that the actual transition caused cannot generally be predicted with certainty. The transition function $T : \mathcal{S} \times \mathcal{A} \to \Delta(\mathcal{S})$ describes the effects of each action at each state. $T(s_i, a)$ is a probability distribution over $\mathcal{S}$; specifically, $T(s_i, a)(s_j)$ is the probability of ending up in state $s_j \in \mathcal{S}$ when action $a$ is performed at state $s_i$. We will denote this quantity by $\Pr(s_i, a, s_j)$. We require that $0 \leq \Pr(s_i, a, s_j) \leq 1$ for all $s_i, s_j$, and that for all $s_i$, $\sum_{s_j \in \mathcal{S}} \Pr(s_i, a, s_j) = 1$. The components $\mathcal{S}$, $\mathcal{A}$ and $T$ determine the dynamics of the system being controlled. The assumption that the system is fully observable means that the agent knows the true state at each time $t$ (once that stage is reached), and its decisions can be based solely on this knowledge. Thus, uncertainty lies only in the prediction of an action's effects, not in determining its actual effect after its execution.

A *(deterministic, stationary, Markovian) policy* $\pi : \mathcal{S} \to \mathcal{A}$ describes a course of action to be adopted by an agent controlling the system. An agent adopting such a policy performs action $\pi(s)$ whenever it finds itself in state $s$. Policies of this form are Markovian since the action choice at any state does not depend on the system history, and are stationary since action choice does not depend on the stage of the decision problem. For the problems we consider, optimal stationary Markovian policies always exist.

We assume a bounded, real-valued reward function $R : \mathcal{S} \to \Re$. $R(s)$ is the instantaneous reward an agent receives for occupying state $s$. A number of optimality criteria can be adopted to measure the value of a policy $\pi$, all measuring in some way the reward accumulated by an agent as it traverses the state space through the execution of $\pi$. In this work, we focus on *discounted infinite-horizon* problems: the *current* value of a reward received $t$ stages in the future is discounted by some factor $\gamma^t (0 \leq \gamma < 1)$. This allows simpler computational methods to be used, as discounted total reward will be finite. Discounting can be justified on other (e.g., economic) grounds in many situations as well.

The *value function* $V_\pi : S \to \Re$ reflects the value of a policy $\pi$ at any state $s$; this is simply the expected sum of discounted future rewards obtained by executing $\pi$ beginning at $s$. A policy $\pi^*$ is *optimal* if, for all $s \in \mathcal{S}$ and all policies $\pi$, we have $V_{\pi^*}(s) \geq V_\pi(s)$. We are guaranteed that such optimal (stationary) policies exist in our setting (Puterman, 1994). The (optimal) value of a state $V^*(s)$ is its value $V_{\pi^*}(s)$ under any optimal policy $\pi^*$.

By *solving* an MDP, we refer to the problem of constructing an optimal policy. *Value iteration* (Bellman, 1957) is a simple iterative approximation algorithm for optimal policy construction. Given some arbitrary estimate $V^0$ of the true value function $V^*$, we iteratively improve this estimate as follows:

$$V^n(s_i) = R(s_i) + \max_{a \in \mathcal{A}} \{ \gamma \sum_{s_j \in \mathcal{S}} \Pr(s_i, a, s_j) V^{n-1}(s_j) \} \tag{1}$$

The computation of $V^n(s)$ given $V^{n-1}$ is known as a *Bellman backup*. The sequence of value functions $V^n$ produced by value iteration converges linearly to $V^*$. Each iteration of value iteration requires $O(|\mathcal{S}|^2 |\mathcal{A}|)$ computation time, and the number of iterations is polynomial in $|\mathcal{S}|$.

For some finite $n$, the actions $a$ that maximize the right-hand side of Equation 1 form an optimal policy, and $V^n$ approximates its value. Various termination criteria can be applied; for example, one might terminate the algorithm when

$$\|V^{i+1} - V^i\| \leq \frac{\varepsilon(1-\gamma)}{2\gamma} \tag{2}$$

(where $\|X\| = \max\{|x| : x \in X\}$ denotes the supremum norm). This ensures the resulting value function $V^{i+1}$ is within $\frac{\varepsilon}{2}$ of the optimal function $V^*$ at any state, and that the induced policy is $\varepsilon$-optimal (i.e., its value is within $\varepsilon$ of $V^*$) (Puterman, 1994).

A concept that will be useful later is that of a *Q-function*. Given an arbitrary value function $V$, we define $Q_a^V(s_i)$ as

$$Q_a^V(s_i) = R(s_i) + \gamma \sum_{s_j \in \mathcal{S}} \Pr(s_i, a, s_j) V(s_j) \tag{3}$$

Intuitively, $Q_a^V(s)$ denotes the value of performing action $a$ at state $s$ and then acting in a manner that has value $V$ (Watkins & Dayan, 1992). In particular, we define $Q_a^*$ to be the Q-function defined with respect to $V^*$, and $Q_a^n$ to be the Q-function defined with respect to $V^{n-1}$. In this manner, we can rewrite Equation 1 as:

$$V^n(s) = \max_{a \in \mathcal{A}} \{ Q_a^n(s) \} \tag{4}$$

We define an ergodic MDP as an MDP in which every state is reachable from any other state in a finite number of steps with non-zero probability.

## 2.2 Model-based Reinforcement Learning

One difficulty with the use of MDPs is that the construction of an optimal policy requires that the agent know the exact transition probabilities Pr and reward model $R$. In the specification of a decision problem, these requirements, especially the detailed specification of the domain's dynamics, can impose an undue burden on the agent's designer. *Reinforcement learning* can be viewed as solving an MDP in which the full details of the model, in particular Pr and $R$, are not known to the agent. Instead, the agent learns how to act optimally through experience with its environment. We provide a brief overview of reinforcement learning in this section (with an emphasis on model-based approaches). For further details, please refer to the texts of Sutton and Barto (1998) and Bertsekas and Tsitsiklis (1996), and the survey of Kaelbling, Littman and Moore (1996).

In the general model, we assume that an agent is controlling an MDP $\langle \mathcal{S}, \mathcal{A}, T, R \rangle$ and initially knows its state and action spaces, $\mathcal{S}$ and $\mathcal{A}$, but not the transition model $T$ or reward function $R$. The agent acts in its environment, and at each stage of the process makes a "transition" $\langle s, a, r, t \rangle$; that is, it takes action $a$ at state $s$, receives reward $r$ and moves to state $t$. Based on repeated experiences of this type it can determine an optimal policy in one of two ways: (a) in *model-based reinforcement learning*, these experiences can be used to learn the true nature of $T$ and $R$, and the MDP can be solved using standard methods (e.g., value iteration); or (b) in *model-free reinforcement learning*, these experiences can be used to directly update an estimate of the optimal value function or Q-function.

Probably the simplest model-based reinforcement learning scheme is the *certainty equivalence approach*. Intuitively, a learning agent is assumed to have some current estimated transition model $\widehat{T}$ of its environment consisting of estimated probabilities $\widehat{\Pr}(s, a, t)$ and an estimated rewards model $\widehat{R}(s)$. With each experience $\langle s, a, r, t \rangle$ the agent updates its estimated models, solves the estimated MDP $\widehat{M}$ to obtain an policy $\widehat{\pi}$ that would be optimal if its estimated models were correct, and acts according to that policy.

To make the certainty equivalence approach precise, a specific form of estimated model and update procedure must be adopted. A common approach is to used the empirical distribution of observed state transitions and rewards as the estimated model. For instance, if action $a$ has been attempted $C(s, a)$ times at state $s$, and on $C(s, a, t)$ of those occasions state $t$ has been reached, then the estimate $\widehat{\Pr}(s, a, t) = C(s, a, t)/C(s, a)$. If $C(s, a) = 0$, some prior estimate is used (e.g., one might assume all state transitions are equiprobable). A Bayesian approach (Dearden, Friedman, & Andre, 1999) uses an explicit prior distribution over the parameters of the transition distribution $\Pr(s, a, \cdot)$, and then updates these with each experienced transition. For instance, we might assume a Dirichlet (Generalized Beta) distribution (DeGroot, 1975) with parameters $n(s, a, t)$ associated with each possible successor state $t$. The Dirichlet parameters are equal to the experience-based counts $C(s, a, t)$ plus a "prior count" $P(s, a, t)$ representing the agent's prior beliefs about the distribution (i.e., $n(s, a, t) = C(s, a, t) + P(s, a, t)$). The *expected* transition probability $\Pr(s, a, t)$ is then $n(s, a, t)/\sum_{t'} n(s, a, t')$. Assuming parameter independence, the MDP $\widehat{M}$ can be solved using these expected values. Furthermore, the model can be updated with ease, simply increasing $n(s, a, t)$ by one with each observation $\langle s, a, r, t \rangle$. This model has the advantage over a counter-based approach of allowing a flexible prior model and generally does not assign probability zero to unobserved transitions. We will adopt this Bayesian perspective in our imitation model.

One difficulty with the certainty equivalence approach is the computational burden of re-solving an MDP $\widehat{M}$ with each update of the models $\widehat{T}$ and $\widehat{R}$ (i.e., with each experience). One could circumvent this to some extent by batching experiences and updating (and re-solving) the model only periodically. Alternatively, one could use computational effort judiciously to apply Bellman backups only at those states whose values (or Q-values) are likely to change the most given a change in the model. Moore and Atkeson's (1993) *prioritized sweeping* algorithm does just this. When $\widehat{T}$ is updated by changing $\widehat{\Pr}(s, a, t)$, a Bellman backup is applied at $s$ to update its estimated value $\widehat{V}$, as well as the Q-value $\widehat{Q}(s, a)$. Suppose the magnitude of the change in $\widehat{V}(s)$ is given by $\Delta \widehat{V}(s)$. For any predecessor $w$, the Q-values $\widehat{Q}(w, a')$—hence values $\widehat{V}(w)$—can change if $\widehat{\Pr}(w, a', s) > 0$. The magnitude of the change is bounded by $\widehat{\Pr}(w, a', s) \Delta \widehat{V}(s)$. All such predecessors $w$ of $s$ are placed in a priority queue with $\widehat{\Pr}(w, a', s) \Delta \widehat{V}(s)$ serving as the priority. A fixed number of Bellman backups are applied to states in the order in which they appear in the queue. With each backup, any change in value can cause new predecessors to be inserted into the queue. In this way, computational effort is focused on those states where a Bellman backup has the greatest impact due to the model change. Furthermore, the backups are applied only to a subset of states, and are generally only applied a fixed number of times. By way of contrast, in the certainty equivalence approach, backups are applied until convergence. Thus prioritized sweeping can be viewed as a specific form of asynchronous value iteration, and has appealing computational properties (Moore & Atkeson, 1993).

Under certainty equivalence, the agent acts as if the current approximation of the model is correct, even though the model is likely to be inaccurate early in the learning process. If the optimal policy for this *inaccurate* model prevents the agent from exploring the transitions which form part of the optimal policy for the true model, then the agent will fail to find the optimal policy. For this reason, explicit *exploration policies* are invariably used to ensure that each action is tried at each state sufficiently often. By acting randomly (assuming an ergodic MDP), an agent is assured of sampling each action at each state infinitely often in the limit. Unfortunately, the actions of such an agent will fail to exploit (in fact, will be completely uninfluenced by) its knowledge of the optimal policy. This *exploration-exploitation tradeoff* refers to the tension between trying new actions in order to find out more about the environment and executing actions believed to be optimal on the basis of the current estimated model.

The most common method for exploration is the $\varepsilon$–greedy method in which the agent chooses a random action a fraction $\varepsilon$ of the time, where $0 < \varepsilon < 1$. Typically, $\varepsilon$ is decayed over time to increase the agent's exploitation of its knowledge. In the Boltzmann approach, each action is selected with a probability proportional to its value:

$$\Pr_s(a) = \frac{e^{Q(s,a)/\tau}}{\sum_{a' \in \mathcal{A}} e^{Q(s,a')/\tau}} \tag{5}$$

The proportionality can be adjusted nonlinearly with the temperature parameter $\tau$. As $\tau \to 0$ the probability of selecting the action with the highest value tends to 1. Typically, $\tau$ is started high so that actions are randomly explored during the early stages of learning. As the agent gains knowledge about the effects of its actions and the value of these effects, the parameter $\tau$ is decayed so that the agent spends more time exploiting actions known to be valuable and less time randomly exploring actions.

More sophisticated methods attempt to use information about model confidence and value magnitudes to plan a utility-maximizing exploration plan. An early approximation of this scheme can be found in the interval estimation method (Kaelbling, 1993). Bayesian methods have also been used to calculate the expected value of information to be gained from exploration (Meuleau & Bourgine, 1999; Dearden et al., 1999).

We concentrate in this paper on model-based approaches to reinforcement learning. However, we should point out that *model-free methods*—those in which an estimate of the optimal value function or Q-function is learned directly, without recourse to a domain model—have attracted much attention. For example, TD-methods (Sutton, 1988) and Q-learning (Watkins & Dayan, 1992) have both proven to be among the more popular methods for reinforcement learning. Our methods can be modified to deal with model-free approaches, as we discuss in the concluding section. We also focus on so-called *table-based (or explicit) representations* of models and value functions. When state and action spaces are large, table-based approaches become unwieldy, and the associated algorithms are generally intractable. In these situations, approximators are often used to estimate the values of states. We will discuss ways in which our techniques can be extended to allow for function approximation in the concluding section.

## 3. A Formal Framework for Implicit Imitation

To model the influence that a mentor agent can have on the decision process or the learning behavior of an observer, we must extend the single-agent decision model of MDPs to account for the actions and objectives of multiple agents. In this section, we introduce a formal framework for studying implicit imitation. We begin by introducing a general model for stochastic games (Shapley, 1953; Myerson, 1991), and then impose various assumptions and restrictions on this general model that allow us to focus on the key aspects of implicit imitation. We note that the framework proposed here is useful for the study of other forms of knowledge transfer in multiagent systems, and we briefly point out various extensions of the framework that would permit implicit imitation, and other forms of knowledge transfer, in more general settings.

### 3.1 Non-Interacting Stochastic Games

*Stochastic games* can be viewed as a multiagent extension of Markov decision processes. Though Shapley's (1953) original formulation of stochastic games involved a zero-sum (fully competitive) assumption, various generalizations of the model have been proposed allowing for arbitrary relationships between agents' utility functions (Myerson, 1991).[2] Formally, an *n*-agent stochastic game $\langle \mathcal{S}, \{\mathcal{A}_i : i \leq n\}, T, \{R_i : i \leq n\}\rangle$ comprises a set of $n$ agents ($1 \leq i \leq n$), a set of states $\mathcal{S}$, a set of actions $\mathcal{A}_i$ for each agent $i$, a state transition function $T$, and a reward function $R_i$ for each agent $i$. Unlike an MDP, individual agent actions do not determine state transitions; rather it is the *joint action* taken by the collection of agents that determines how the system evolves at any point in time. Let $\mathcal{A} = \mathcal{A}_1 \times \cdots \times \mathcal{A}_n$ be the set of joint actions; then $T : \mathcal{S} \times \mathcal{A} \to \Delta(\mathcal{S})$, with $T(s_i, a)(s_j) = \Pr(s_i, a, s_j)$ denoting the probability of ending up in state $s_j \in \mathcal{S}$ when joint action $a$ is performed at state $s_i$.

---

2. For example, see the fully cooperative multiagent MDP model proposed by Boutilier (1999).

For convenience, we introduce the notation $\mathcal{A}_{-i}$ to denote the set of joint actions $\mathcal{A}_1 \times \cdots \times \mathcal{A}_{i-1} \times \mathcal{A}_{i+1} \times \cdots \times \mathcal{A}_n$ involving all agents except $i$. We use $a_i \cdot a_{-i}$ to denote the (full) joint action obtained by conjoining $a_i \in \mathcal{A}_i$ with $a_{-i} \in \mathcal{A}_{-i}$.

Because the interests of the individual agents may be at odds, strategic reasoning and notions of equilibrium are generally involved in the solution of stochastic games. Because our aim is to study how a reinforcement agent might learn by observing the behavior of an expert mentor, we wish to restrict the model in such a way that strategic interactions need not be considered: we want to focus on settings in which the actions of the observer and the mentor do not interact. Furthermore, we want to assume that the reward functions of the agents do not conflict in a way that requires strategic reasoning.

We define *noninteracting stochastic games* by appealing to the notion of an agent *projection function* which is used to extract an agent's *local state* from the underlying game. In these games, an agent's local state determines all aspects of the global state that are relevant to its decision making process, while the projection function determines which global states are identical from an agent's local perspective. Formally, for each agent $i$, we assume a local state space $\mathcal{S}_i$, and a projection function $L_i : \mathcal{S} \to \mathcal{S}_i$. For any $s, t \in \mathcal{S}$, we write $s \sim_i t$ iff $L_i(s) = L_i(t)$. This equivalence relation partitions $\mathcal{S}$ into a set of equivalence classes such that the elements within a specific class (i.e., $L_i^{-1}(s)$ for some $s \in \mathcal{S}_i$) need not be distinguished by agent $i$ for the purposes of individual decision making. We say a stochastic game is *noninteracting* if there exists a local state space $\mathcal{S}_i$ and projection function $L_i$ for each agent $i$ such that:

1. If $s \sim_i t$, then $\forall a_i \in \mathcal{A}_i, a_{-i} \in \mathcal{A}_{-i}, w_i \in \mathcal{S}_i$ we have

$$\sum \{\Pr(s, a_i \cdot a_{-i}, w) : w \in L_i^{-1}(w_i)\} = \sum \{\Pr(t, a_i \cdot a_{-i}, w) : w \in L_i^{-1}(w_i)\}$$

2. $R_i(s) = R_i(t)$ if $s \sim_i t$

Intuitively, condition 1 above imposes two distinct requirements on the game from the perspective of agent $i$. First, if we ignore the existence of other agents, it provides a notion of state space abstraction suitable for agent $i$. Specifically, $L_i$ clusters together states $s \in \mathcal{S}$ only if each state in an equivalence class has identical dynamics with respect to the abstraction induced by $L_i$. This type of abstraction is a form of bisimulation of the type studied in automaton minimization (Hartmanis & Stearns, 1966; Lee & Yannakakis, 1992) and automatic abstraction methods developed for MDPs (Dearden & Boutilier, 1997; Dean & Givan, 1997). It is not hard to show—ignoring the presence of other agents—that the underlying system is Markovian with respect to the abstraction (or equivalently, w.r.t. $\mathcal{S}_i$) if condition 1 is met. The quantification over all $a_{-i}$ imposes a strong noninteraction requirement, namely, that the dynamics of the game from the perspective of agent $i$ is independent of the strategies of the other agents. Condition 2 simply requires that all states within a given equivalence class for agent $i$ have the same reward for agent $i$. This means that no states within a class need to be distinguished—each local state can be viewed as atomic.

A noninteracting game induces an MDP $M_i$ for each agent $i$ where $M_i = \langle \mathcal{S}_i, \mathcal{A}_i, \mathrm{Pr}_i, R_i \rangle$ where $\mathrm{Pr}_i$ is given by condition (1) above. Specifically, for each $s_i, t_i \in \mathcal{S}_i$:

$$\mathrm{Pr}_i(s_i, a_i, t_i) = \sum \{\mathrm{Pr}(s, a_i.a_{-i}, t) : t \in L_i^{-1}(t_i)\}$$

where $s$ is *any* state in $L_i^{-1}(s_i)$ and $a_{-i}$ is *any* element of $\mathcal{A}_{-i}$. Let $\pi_i : \mathcal{S}_a \to \mathcal{A}_i$ be an optimal policy for $M_i$. We can extend this to a strategy $\pi_i^G : \mathcal{S} \to \mathcal{A}_i$ for the underlying stochastic game by simply applying $\pi_i(s_i)$ to every state $s \in \mathcal{S}$ such that $L_i(s) = s_i$. The following proposition shows that the term "noninteracting" indeed provides an appropriate description of such a game.

**Proposition 1** *Let $G$ be a noninteracting stochastic game, $M_i$ the induced MDP for agent $i$, and $\pi_i$ some optimal policy for $M_i$. The strategy $\pi_i^G$ extending $\pi_i$ to $G$ is dominant for agent $i$.*

Thus each agent can solve the noninteracting game by abstracting away irrelevant aspects of the state space, ignoring other agent actions, and solving its "personal" MDP $M_i$.

Given an arbitrary stochastic game, it can generally be quite difficult to discover whether it is noninteracting, requiring the construction of appropriate projection functions. In what follows, we will simply *assume* that the underlying multiagent system is a noninteracting game. Rather than specifying the game and projection functions, we will specify the individual MDPs $M_i$ themselves. The noninteracting game induced by the set of individual MDPs is simply the "cross product" of the individual MDPs. Such a view is often quite natural. Consider the example of three robots moving in some two-dimensional office domain. If we are able to neglect the possibility of interaction—for example, if the robots can occupy the same 2-D position (at a suitable level of granularity) and do not require the same resources to achieve their tasks—then we might specify an individual MDP for each robot. The local state might be determined by the robot's $x, y$-position, orientation, and the status of its own tasks. The global state space would be the cross product $\mathcal{S}_1 \times \mathcal{S}_2 \times \mathcal{S}_3$ of the local spaces. The individual components of any joint action would affect only the local state, and each agent would care (through its reward function $R_i$) only about its local state.

We note that the projection function $L_i$ should not be viewed as equivalent to an observation function. We do not assume that agent $i$ can only distinguish elements of $S_i$—in fact, observations of other agents' states will be crucial for imitation. Rather the existence of $L_i$ simply means that, *from the point of view of decision making with a known model*, the agent need not worry about distinctions other than those made by $L_i$. Assuming no computational limitations, an agent $i$ need only solve $M_i$, but may use observations of other agents in order to improve its knowledge about $M_i$'s dynamics.[3]

## 3.2 Implicit Imitation

Despite the very independent nature of the agent subprocesses in a noninteracting multiagent system, there are circumstances in which the behavior of one agent may be relevant to another. To keep the discussion simple, we assume the existence of an expert *mentor* agent $m$, which is implementing some stationary (and presumably optimal) policy $\pi_m$ over its local MDP $M_m = \langle \mathcal{S}_m, \mathcal{A}_m, \text{Pr}_m, R_m \rangle$. We also assume a second agent $o$, the *observer*, with local MDP $M_o = \langle \mathcal{S}_o, \mathcal{A}_o, \text{Pr}_o, R_o \rangle$. While nothing about the mentor's behavior is relevant to the observer if it knows its own MDP (and can solve it without computational difficulty), the situation can be quite different if $o$ is a reinforcement learner without complete knowledge of the model $M_o$. It may well be that the observed behavior of the mentor provides valuable information to the observer in its quest to learn how to act optimally within $M_o$. To take an extreme case, if mentor's MDP is identical to the observer's, and the mentor is an expert (in the sense of acting optimally), then the behavior of the mentor indicates exactly what the observer should do. Even if the mentor is not acting optimally, or if the mentor and observer have different reward functions, mentor state transitions observed by the learner can provide valuable information about the dynamics of the domain.

---

3. We elaborate on the condition of computational limitations below.

Thus we see that when one agent is learning how to act, the behavior of another can potentially be relevant to the learner, even if the underlying multiagent system is noninteracting. Similar remarks, of course, apply to the case where the observer knows the MDP $M_o$, but computational restrictions make solving this difficult—observed mentor transitions might provide valuable information about where to focus computational effort.[4] The main motivation underlying our model of implicit imitation is that the behavior of an expert mentor can provide hints as to appropriate courses of action for a reinforcement learning agent.

Intuitively, *implicit imitation* is a mechanism by which a learning agent attempts to incorporate the observed experience of an expert mentor agent into its learning process. Like more classical forms of learning by imitation, the learner considers the effects of the mentor's action (or action sequence) in its own context. Unlike direct imitation, however, we do not assume that the learner must "physically" attempt to duplicate the mentor's behavior, nor do we assume that the mentor's behavior is necessarily appropriate for the observer. Instead, the influence of the mentor is on the agent's *transition model* and its estimate of *value* of various states and actions. We elaborate on these points below.

In what follows, we assume a mentor $m$ and associated MDP $M_m$, and a learner or observer $o$ and associated MDP $M_o$, as described above. These MDPs are fully observable. We focus on the reinforcement learning problem faced by agent $o$. The extension to multiple mentors is straightforward and will be discussed below, but for clarity we assume only one mentor in our description of the abstract framework. It is clear that certain conditions must be met for the observer to extract useful information from the mentor. We list a number of assumptions that we make at different points in the development of our model.

**Observability:** We must assume that the learner can *observe* certain aspects of the mentor's behavior. In this work, we assume that state of the mentor's MDP is *fully observable* to the learner. Equivalently, we interpret this as full observability of the underlying noninteracting game, together with knowledge of the mentor's projection function $L_m$. A more general partially observable model would require the specification of an observation or signal set $\mathcal{Z}$ and an observation function $O : \mathcal{S}_o \times \mathcal{S}_m \to \Delta(\mathcal{Z})$, where $O(s_o, s_m)(z)$ denotes the probability with which the observer obtains signal $z$ when the local states of the observer and mentor are $s_o$ and $s_m$, respectively. We do not pursue such a model here. It is important to note that we do *not* assume that the observer has access to the *action* taken by $m$ at any point in time. Since actions are stochastic, the state (even if fully observable) that results from the mentor invoking a specific control signal is generally insufficient to determine that signal. Thus it seems much more reasonable to assume that states (and transitions) are observable than the actions that gave rise to them.

---

4. For instance, algorithms like asynchronous dynamic programming and prioritized sweeping can benefit from such guidance. Indeed, the distinction between reinforcement learning and solving MDPs is viewed by some as rather blurry (Sutton & Barto, 1998; Bertsekas & Tsitsiklis, 1996). Our focus is on the case of an unknown model (i.e., the classical reinforcement learning problem) as opposed to one where computational issues are key.

**Analogy:** If the observer and the mentor are acting in different local state spaces, it is clear that observations made of the mentor's state transitions can offer no useful information to the observer unless there is some relationship between the two state spaces. There are several ways in which this relationship can be specified. Dautenhahn and Nehaniv (1998) use a homomorphism to define the relationship between mentor and observer for a *specific* family of trajectories (see Section 8 for further discussion).

A slightly different notion might involve the use of some analogical mapping $h : \mathcal{S}_m \to \mathcal{S}_o$ such that an observed state transition $s \to t$ provides some information to the observer about the dynamics or value of state $h(s) \in \mathcal{S}_o$. In certain circumstances, we might require the mapping $h$ to be homomorphic with respect to $\Pr(\cdot, a, \cdot)$ (for some, or all, $a$), and perhaps even with respect to $R$. We discuss these issues in further detail below. In order to simplify our model and avoid undue attention to the (admittedly important) topic of constructing suitable analogical mappings, we will simply assume that the mentor and the observer have "identical" state spaces; that is, $\mathcal{S}_m$ and $\mathcal{S}_o$ are in some sense isomorphic. The precise sense in which the spaces are isomorphic—or in some cases, *presumed* to be isomorphic until proven otherwise—is elaborated below when we discuss the relationship between agent abilities. Thus from this point we simply refer to the state $\mathcal{S}$ without distinguishing the mentor's local space $\mathcal{S}_m$ from the observer's $\mathcal{S}_o$.

**Abilities:** Even with a mapping between states, observations of a mentor's state transitions only tell the observer something about the mentor's abilities, not its own. We must assume that the observer can in some way "duplicate" the actions taken by the mentor to induce analogous transitions in its own local state space. In other words, there must be some presumption that the mentor and the observer have similar abilities. It is in this sense that the analogical mapping between state spaces can be taken to be a homomorphism. Specifically, we might assume that the mentor and the observer have the same actions available to them (i.e., $\mathcal{A}_m = \mathcal{A}_o = \mathcal{A}$) and that $h : \mathcal{S}_m \to \mathcal{S}_o$ is homomorphic with respect to $\Pr(\cdot, a, \cdot)$ for all $a \in \mathcal{A}$. This requirement can be weakened substantially, without diminishing its utility, by requiring only that the observer be able to implement the actions *actually taken* by the mentor at a given state $s$. Finally, we might have an observer that *assumes* that it can duplicate the actions taken by the mentor until it finds evidence to the contrary. In this case, there is a *presumed homomorphism* between the state spaces. In what follows, we will distinguish between implicit imitation in *homogeneous action settings*—domains in which the analogical mapping is indeed homomorphic—and *heterogeneous action settings*—where the mapping may not be a homomorphism.

There are more general ways of defining similarity of ability, for example, by assuming that the observer may be able to move through state space in a similar fashion to the mentor without following the same trajectories (Nehaniv & Dautenhahn, 1998). For instance, the mentor may have a way of moving directly between key locations in state space, while the observer may be able to move between analogous locations in a less direct fashion. In such a case, the analogy between states may not be determined by single actions, but rather by sequences of actions or local policies. We will suggest ways for dealing with restricted forms of analogy of this type in Section 5.

**Objectives:** Even when the observer and mentor have similar or identical abilities, the value to the observer of the information gleaned from the mentor may depend on the actual policy being implemented by the mentor. We might suppose that the more closely related a mentor's policy is to the optimal policy of the observer, the more useful the information will be. Thus, to some extent, we expect that the more closely aligned the objectives of the mentor and the observer are, the more valuable the guidance provided by the mentor. Unlike in existing teaching models, we do not suppose that the mentor is making any explicit efforts to instruct the observer. And because their objectives may not be identical, we do not force the observer to (attempt to) explicitly imitate the behavior of the mentor. In general, we will make no explicit assumptions about the relationship between the objectives of the mentor and the observer. However, we will see that, to some extent, the "closer" they are, the more utility can be derived from implicit imitation.

Finally, we remark on an important assumption we make throughout the remainder of this paper: the observer knows its reward function $R_o$; that is, for each state $s$, the observer can evaluate $R_o(s)$ without having visited state $s$. This view is consistent with view of reinforcement learning as "automatic programming." A user may easily specify a reward function (e.g., in the form of a set of predicates that can be evaluated at any state) prior to learning. It may be more difficult to specify a domain model or optimal policy. In such a setting, the only unknown component of the MDP $M_o$ is the transition function $\Pr_o$. We believe this approach to reinforcement learning is, in fact, more common in practice than the approach in which the reward function must be sampled.

To reiterate, our aim is to describe a mechanism by which the observer can accelerate its learning; but we emphasize our position that *implicit* imitation—in contrast to *explicit* imitation—is not merely replicating the behaviors (or state trajectories) observed in another agent, nor even attempting to reach "similar states". We believe the agent must learn about its own capabilities and adapt the information contained in observed behavior to these. Agents must also explore the appropriate application (if any) of observed behaviors, integrating these with their own, as appropriate, to achieve their own ends. We therefore see imitation as an interactive process in which the behavior of one agent is used to guide the learning of another.

Given this setting, we can list possible ways in which an observer and a mentor can (and cannot) interact, contrasting along the way our perspective and assumptions with those of existing models in the literature.[5] First, the observer could attempt to directly infer a policy from its observations of mentor state-action pairs. This model has a conceptual simplicity and intuitive appeal, and forms the basis of the behavioral cloning paradigm (Sammut, Hurst, Kedzier, & Michie, 1992; Urbancic & Bratko, 1994). However, it assumes that the observer and mentor share the same reward function and action capabilities. It also assumes that complete and unambiguous trajectories (*including* action choices) can be observed. A related approach attempts to deduce constraints on the value function from the inferred action preferences of the mentor agent (Utgoff & Clouse, 1991; Šuc & Bratko, 1997). Again, however, this approach assumes congruity of objectives. Our model is also distinct from models of explicit teaching (Lin, 1992; Whitehead, 1991b): we do not assume that the mentor has any incentive to move through its environment in a way that explicitly guides the learner to explore its own environment and action space more effectively.

Instead of trying to directly learn a policy, an observer could attempt to use observed state transitions of other agents to improve its *own* environment model $\Pr_o(s, a, t)$. With a more accurate model and its own reward function, the observer could calculate more accurate values for states. The state values could then be used to guide the agent towards distant rewards and reduce the need for random exploration. This insight forms the core of our implicit imitation model. This approach has not been developed in the literature, and is appropriate under the conditions listed above, specifically, under conditions where the mentor's actions are unobservable, and the mentor and observer have different reward functions or objectives. Thus, this approach is applicable under more general conditions than many existing models of imitation learning and teaching.

In addition to model information, mentors may also communicate information about the relevance or irrelevance of regions of the state space for certain classes of reward functions. An observer can use the set of states visited by the mentor as heuristic guidance about where to perform backup computations in the state space.

In the next two sections, we develop specific algorithms from our insights about how agents can use observations of others to both improve their own models and assess the relevance of regions within their state spaces. We first focus on the homogeneous action case, then extend the model to deal with heterogeneous actions.

## 4. Implicit Imitation in Homogeneous Settings

We begin by describing implicit imitation in homogeneous action settings—the extension to heterogeneous settings will build on the insights developed in this section. We develop a technique called *implicit imitation* through which observations of a mentor can be used to accelerate reinforcement learning. First, we define the homogeneous setting. Then we develop the implicit imitation algorithm. Finally, we demonstrate how implicit imitation works on a number of simple problems designed to illustrate the role of the various mechanisms we describe.

---

5. We will describe other models in more detail in Section 8.

## 4.1 Homogeneous Actions

The homogeneous action setting is defined as follows. We assume a single mentor $m$ and observer $o$, with individual MDPs $M_m = \langle \mathcal{S}, \mathcal{A}_m, \Pr_m, R_m \rangle$ and $M_o = \langle \mathcal{S}, \mathcal{A}_o, \Pr_o, R_o \rangle$, respectively. Note that the agents share the same state space (more precisely, we assume a trivial isomorphic mapping that allows us to identify their local states). We also assume that the mentor is executing some stationary policy $\pi_m$. We will often treat this policy as deterministic, but most of our remarks apply to stochastic policies as well. Let the *support set $Supp(\pi_m, s)$* for $\pi_m$ at state $s$ be the set of actions $a \in \mathcal{A}_m$ accorded nonzero probability by $\pi_m$ at state $s$. We assume that the observer has the same abilities as the mentor in the following sense: $\forall s, t \in \mathcal{S}, a_m \in Supp(\pi_m, s)$, there exists an action $a_o \in \mathcal{A}_o$ such that $\Pr_o(s, a_o, t) = \Pr_m(s, a_m, t)$. In other words, the observer is able to duplicate (in a the sense of inducing the same distribution over successor states) the *actual behavior* of the mentor; or equivalently, the agents' local state spaces are isomorphic with respect to the actions actually taken by the mentor at the subset of states where those actions might be taken. This is much weaker than requiring a full homomorphism from $\mathcal{S}_m$ to $\mathcal{S}_o$. Of course, the existence of a full homomorphism is sufficient from our perspective; but our results do not require this.

## 4.2 The Implicit Imitation Algorithm

The implicit imitation algorithm can be understood in terms of its component processes. First, we extract action models from a mentor. Then we integrate this information into the observer's own value estimates by augmenting the usual Bellman backup with mentor action models. A confidence testing procedure ensures that we only use this augmented model when the observer's model of the mentor is more reliable than the observer's model of its own behavior. We also extract occupancy information from the observations of mentor trajectories in order to focus the observer's computational effort (to some extent) in specific parts of the state space. Finally, we augment our action selection process to choose actions that will explore high-value regions revealed by the mentor. The remainder of this section expands upon each of these processes and how they fit together.

### 4.2.1 Model Extraction

The information available to the observer in its quest to learn how to act optimally can be divided into two categories. First, with each action it takes, it receives an experience tuple $\langle s, a, r, t \rangle$; in fact, we will often ignore the sampled reward $r$, since we assume the reward function $R$ is known in advance. As in standard model-based learning, each such experience can be used to update its own transition model $\Pr_o(s, a, \cdot)$.

Second, with each mentor transition, the observer obtains an experience tuple $\langle s, t \rangle$. Note again that the observer does not have direct access to the action taken by the mentor, only the induced state transition. Assume the mentor is implementing a deterministic, stationary policy $\pi_m$, with $\pi_m(s)$ denoting the mentor's choice of action at state $s$. This policy induces a Markov chain $\Pr_m(\cdot, \cdot)$ over $\mathcal{S}$, with $\Pr_m(s, t) = \Pr(s, \pi_m(s), t)$ denoting the probability of a transition from $s$ to $t$.[6] Since the learner observes the mentor's state transitions, it can construct an estimate $\widehat{\Pr}_m$ of this chain: $\widehat{\Pr}_m(s,t)$ is simply estimated by the relative observed frequency of mentor transitions $s \to t$ (w.r.t. all transitions taken from $s$). If the observer has some prior over the possible mentor transitions, standard Bayesian update techniques can be used instead. We use the term *model extraction* for this process of estimating the mentor's Markov chain.

### 4.2.2 AUGMENTED BELLMAN BACKUPS

Suppose the observer has constructed an estimate $\widehat{\Pr}_m$ of the mentor's Markov chain. By the homogeneity assumption, the action $\pi_m(s)$ can be replicated exactly by the observer at state $s$. Thus, the policy $\pi_m$ can, in principle, be duplicated by the observer (were it able to identify the actual actions used). As such, we can define the value of the mentor's policy *from the observer's perspective*:

$$V_m(s) = R_o(s) + \gamma \sum_{t \in S} \Pr_m(s,t) V_m(t) \qquad (6)$$

Notice that Equation 6 uses the mentor's dynamics but the observer's reward function. Letting $V$ denote the optimal (observer's) value function, clearly $V(s) \geq V_m(s)$, so $V_m$ provides a lower bound on the observer's value function.

More importantly, the terms making up $V_m(s)$ can be integrated directly into the Bellman equation for the observer's MDP, forming the *augmented Bellman equation*:

$$V(s) = R_o(s) + \gamma \max \left\{ \max_{a \in A_o} \left\{ \sum_{t \in S} \Pr_o(s,a,t) V(t) \right\}, \sum_{t \in S} \Pr_m(s,t) V(t) \right\} \qquad (7)$$

This is the usual Bellman equation with an extra term added, namely, the second summation, $\sum_{t \in S} \Pr_m(s,t) V(t)$ denoting the expected value of duplicating the mentor's action $a_m$. Since this (unknown) action is identical to one of the observer's actions, the term is redundant and the augmented value equation is valid. Of course, the observer using the augmented backup operation must rely on estimates of these quantities. If the observer exploration policy ensures that each state is visited infinitely often, the estimates of the $\Pr_o$ terms will converge to their true values. If the mentor's policy is ergodic over state space $S$, then $\Pr_m$ will also converge to its true value. If the mentor's policy is restricted to a subset of states $S' \subseteq S$ (those forming the basis of its Markov chain), then the estimates of $\Pr_m$ for the subset will converge correctly with respect to $S'$ if the chain is ergodic. The states in $S - S'$ will remain unvisited and the estimates will remain uninformed by data. Since the mentor's policy is not under the control of the observer, there is no way for the observer to influence the distribution of samples attained for $\Pr_m$. An observer must therefore be able to reason about the accuracy of the estimated model $\Pr_m$ for any $s$ and restrict the application of the augmented equation to those states where $\Pr_m$ is known with sufficient accuracy.

---

6. This is somewhat imprecise, since the initial distribution of the Markov chain is unknown. For our purposes, it is only the dynamics that are relevant to the observer, so only the transition probabilities are used.

While $\Pr_m$ cannot be used indiscriminately, we argue that it can be highly informative early in the learning process. Assuming that the mentor is pursuing an optimal policy (or at least is behaving in some way so that it tends to visit certain states more frequently), there will be many states for which the observer has much more accurate estimates of $\Pr_m(s, t)$ than it does for $\Pr_o(s, a, t)$ for any specific $a$. Since the observer is learning, it must explore both its state space—causing less frequent visits to $s$—and its action space—thus spreading its experience at $s$ over all actions $a$. This generally ensures that the sample size upon which $\Pr_m$ is based is greater than that for $\Pr_o$ for any action that forms part of the mentor's policy. Apart from being more accurate, the use of $\Pr_m(s, t)$ can often give more informed value estimates at state $s$, since prior action models are often "flat" or uniform, and only become distinguishable at a given state when the observer has sufficient experience at state $s$.

We note that the reasoning above holds even if the mentor is implementing a (stationary) stochastic policy (since the expected value of stochastic policy for a fully-observable MDP cannot be greater than that of an optimal deterministic policy). While the "direction" offered by a mentor implementing a deterministic policy tends to be more *focused*, empirically we have found that mentors offer *broader* guidance in moderately stochastic environments or when they implement stochastic policies, since they tend to visit more of the state space. We note that the extension to multiple mentors is straightforward—each mentor model can be incorporated into the augmented Bellman equation without difficulty.

### 4.2.3 MODEL CONFIDENCE

When the mentor's Markov chain is not ergodic, or if the mixing rate[7] is sufficiently low, the mentor may visit a certain state $s$ relatively infrequently. The estimated mentor transition model corresponding to a state that is rarely (or never) visited by the mentor may provide a very misleading estimate—based on the small sample or the prior for the mentor's chain—of the value of the mentor's (unknown) action at $s$; and since the mentor's policy is not under the control of the observer, this misleading value may persist for an extended period. Since the augmented Bellman equation does not consider relative reliability of the mentor and observer models, the value of such a state $s$ may be overestimated;[8] that is, the observer can be tricked into overvaluing the mentor's (unknown) action, and consequently overestimating the value of state $s$.

To overcome this, we incorporate an estimate of model confidence into our augmented backups. For both the mentor's Markov chain and the observer's action transitions, we assume a Dirichlet prior over the parameters of each of these multinomial distributions (DeGroot, 1975). These reflect the observer's initial uncertainty about the possible transition probabilities. From sample counts of mentor and observer transitions, we update these distributions. With this information, we could attempt to perform optimal Bayesian estimation of the value function; but when the sample counts are small (and normal approximations are not appropriate), there is no simple, closed form expression for the resultant distributions over values. We could attempt to employ sampling methods, but in the in-terest of simplicity we have employed an approximate method for combining information sources inspired by Kaelbling's (1993) interval estimation method.

---

7. The mixing rate refers to how quickly a Markov chain approaches its stationary distribution.
8. Note that underestimates based on such considerations are not problematic, since the augmented Bellman equation then reduces to the usual Bellman equation.

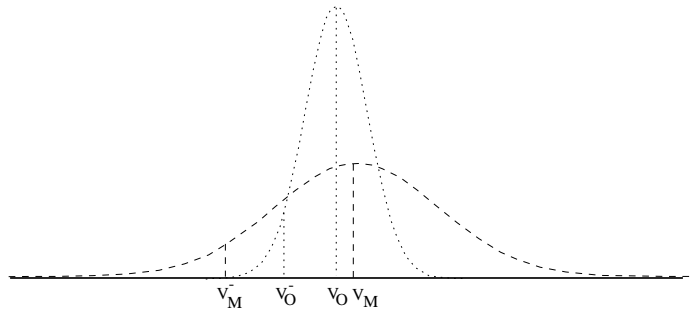

Figure 1: Lower bounds on action values incorporate uncertainty penalty

Let $V$ denote the current estimated augmented value function, and $\Pr_o$ and $\Pr_m$ denote the estimated observer and mentor transition models. We let $\sigma_o^2$ and $\sigma_m^2$ denote the variance in these model parameters.

An augmented Bellman backup with respect to $V$ using confidence testing proceeds as follows. We first compute the observer's optimal action $a_o^*$ based on the estimated augmented values for each of the observer's actions. Let $Q(a_o^*, s) = V_o(s)$ denote its value. For the best action, we use the model uncertainty encoded by the Dirichlet distribution to construct a lower bound $V_o^-(s)$ on the value of the state to the observer using the model (at state $s$) derived from its own behavior (i.e., ignoring its observations of the mentor). We employ transition counts $n_o(s, a, t)$ and $n_m(s, t)$ to denote the number of times the observer has made the transition from state $s$ to state $t$ when the action $a$ was performed, and the number of times the mentor was observed making the transition from state $s$ to $t$, respectively. From these counts, we estimate the uncertainty in the model using the variance of a Dirichlet distribution. Let $\alpha = n_o(s, a, t)$ and $\beta = \sum_{t' \in \mathcal{S} - t} n_o(s, a, t')$. Then the model variance is:

$$\sigma_{\text{model}}^2(s, a, t) = \frac{\alpha + \beta}{(\alpha + \beta)^2 + (\alpha + \beta + 1)} \tag{8}$$

The variance in the Q-value of an action due to the uncertainty in the local model can be found by simple application of the rule for combining linear combinations of variances, $Var(cX + dY) = c^2 Var(X) + d^2 Var(Y)$ to the expression for the Bellman backup, $Var(R(s) + \gamma \sum_t Pr(t|s, a)V(t))$. The result is:

$$\sigma^2(s, a) = \gamma^2 \sum_t \sigma_{\text{model}}^2(s, a, t)v(t)^2 \tag{9}$$

Using Chebychev's inequality,[9] we can obtain a confidence level even though the Dirichlet distributions for small sample counts are highly non-normal. The lower bound is then $V_o^-(s) = V_o(s) - c\sigma_o(s, a_o^*)$ for some suitable constant $c$. One may interpret this as penalizing

---

9. Chebychev's inequality states that $1 - \frac{1}{k^2}$ of the probability mass for an arbitrary distribution will be within $k$ standard deviations of the mean.

FUNCTION augmentedBackup( $V$,$\text{Pr}_o$,$\sigma^2_{omodel}$,$\text{Pr}_m$,$\sigma^2_{mmodel}$,s,c)

$a^* = \arg\max_{a \in \mathcal{A}_o} \sum_{t \in \mathcal{S}} \text{Pr}(s,a,t)V(t)$

$V_o(s) = R_o(s) + \gamma \sum_{t \in \mathcal{S}} \text{Pr}_o(s,a^*,t)V(t)$
$V_m(s) = R_o(s) + \gamma \sum_{t \in \mathcal{S}} \text{Pr}_m(s,t)V(t)$

$\sigma^2_o(s,a^*) = \gamma^2 \sum_{t \in \mathcal{S}} \sigma^2_{omodel}(s,a^*,t)V(t)^2$
$\sigma^2_m(s) = \gamma^2 \sum_{t \in \mathcal{S}} \sigma^2_{mmodel}(s,t)V(t)^2$

$V_o^-(s) = V_o(s) - c * \sigma_o(s,a^*)$
$V_m^-(s) = V_m(s) - c * \sigma_m(s)$

IF $V_o(s)^- > V_m(s)^-$ THEN
$\qquad V(s) = V_o(s)$
ELSE
$\qquad V(s) = V_m(s)$

END

Table 1: Implicit Backup

the value of a state by subtracting its "uncertainty" from it (see Figure 1).[10] The value $V_m(s)$ of the mentor's action $\pi_m(s)$ is estimated similarly and an analogous lower bound $V_m^-(s)$ on it is also constructed. If $V_o^-(s) > V_m^-(s)$, then we say that $V_o(s)$ supersedes $V_m(s)$ and we write $V_o(s) \succ V_m(s)$. When $V_o(s) \succ V_m(s)$ then either the mentor-inspired model has, in fact, a lower expected value (within a specified degree of confidence) and uses a nonoptimal action (from the observer's perspective), or the mentor-inspired model has lower confidence. In either case, we reject the information provided by the mentor and use a standard Bellman backup using the action model derived solely from the observer's experience (thus suppressing the augmented backup)—the backed up value is $V_o(s)$ in this case.

An algorithm for computing an augmented backup using this confidence test is shown in Table 1. The algorithm parameters include the current estimate of the augmented value function $V$, the current estimated model $\text{Pr}_o$ and its associated local variance $\sigma^2_{omodel}$, and the model of the mentor's Markov chain $\text{Pr}_m$ and its associated variance $\sigma^2_{mmodel}$. It calculates lower bounds and returns the mean value, $V_o$ or $V_m$, with the greatest lower bound. The parameter $c$ determines the width of the confidence interval used in the mentor rejection test.

### 4.2.4 FOCUSING

The augmented Bellman backups improves the accuracy of the observer's model. A second way in which an observer can exploit its observations of the mentor is to focus attention on the states visited by the mentor. In a model-based approach, the specific *focusing mecha-*

---

10. Ideally, we would like to take not only the uncertainty of the model at the current state into account, but also the uncertainty of future states as well (Meuleau & Bourgine, 1999).

*nism* we adopt is to require the observer to perform a (possibly augmented) Bellman backup at state $s$ whenever the mentor makes a transition from $s$. This has three effects. First, if the mentor tends to visit interesting regions of space (e.g., if it shares a certain reward structure with the observer), then the significant values backed up from mentor-visited states will bias the observer's exploration towards these regions. Second, computational effort will be concentrated toward parts of state space where the estimated model $\widehat{\Pr}_m(s, t)$ changes, and hence where the estimated value of one of the observer's actions may change. Third, computation is focused where the model is likely to be more accurate (as discussed above).

### 4.2.5 ACTION SELECTION

The integration of exploration techniques in the action selection policy is important for any reinforcement learning algorithm to guarantee convergence. In implicit imitation, it plays a second, crucial role in helping the agent exploit the information extracted from the mentor. Our improved convergence results rely on the greedy quality of the exploration strategy to bias an observer towards the higher-valued trajectories revealed by the mentor.

For expediency, we have adopted the $\varepsilon$-greedy action selection method, using an exploration rate $\varepsilon$ that decays over time. We could easily have employed other semi-greedy methods such as Boltzmann exploration. In the presence of a mentor, greedy action selection becomes more complex. The observer examines its own actions at state $s$ in the usual way and obtains a best action $a_o^*$ which has a corresponding value $V_o(s)$. A value is also calculated for the mentor's action $V_m(s)$. If $V_o(s) \succ V_m(s)$, then the observer's own action model is used and the greedy action is defined exactly as if the mentor were not present. If, however, $V_m(s) \succ V_o(s)$ then we would like to define the greedy action to be the action dictated by the mentor's policy at state $s$. Unfortunately, the observer does not know which action this is, so we define the greedy action to be the observer's action "closest" to the mentor's action according to the observer's current model estimates at $s$. More precisely, the action *most similar* to the mentor's at state $s$, denoted $\kappa_m(s)$, is that whose outcome distribution has minimum Kullback-Leibler divergence from the mentor's action outcome distribution:

$$\kappa_m(s) = \operatorname{argmin}_a \left\{ -\sum_t \Pr_o(s, a, t) \log \Pr_m(s, t) \right\} \qquad (10)$$

The observer's own experience-based action models will be poor early in training, so there is a chance that the closest action computation will select the wrong action. We rely on the exploration policy to ensure that each of the observer's actions is sampled appropriately in the long run.[11]

In our present work we have assumed that the state space is large and that the agent will therefore not be able to completely update the Q-function over the whole space. (The intractability of updating the entire state space is one of the motivations for using imitation techniques). In the absence of information about the state's true values, we would like to bias the value of the states along the mentor's trajectories so that they look worthwhile to explore. We do this by assuming bounds on the reward function and setting the initial Q-values over the entire space below this bound. In our simple examples, rewards are strictly positive so we set the bounds to zero. If mentor trajectories intersect any states valued by the observing agent, backups will cause the states on these trajectories to have a higher value than the surrounding states. This causes the greedy step in the exploration method to prefer actions that lead to mentor-visited states over actions for which the agent has no information.

---

11. If the mentor is executing a stochastic policy, the test based on KL-divergence can mislead the learner.

### 4.2.6 Model Extraction in Specific Reinforcement Learning Algorithms

Model extraction, augmented backups, the focusing mechanism, and our extended notion of the greedy action selection, can be integrated into model-based reinforcement learning algorithms with relative ease. Generically, our *implicit imitation algorithm* requires that: (a) the observer maintain an estimate $\widehat{\mathrm{Pr}}_m(s, t)$ of the Markov chain induced by the mentor's policy—this estimate is updated with every observed transition; and (b) that all backups performed to estimate its value function use the augmented backup (Equation 7) with confidence testing. Of course, these backups are implemented using estimated models $\widehat{\mathrm{Pr}}_o(s, a, t)$ and $\widehat{\mathrm{Pr}}_m(s, t)$. In addition, the focusing mechanism requires that an augmented backup be performed at any state visited by the mentor.

We demonstrate the generality of these mechanisms by combining them with the well-known and efficient *prioritized sweeping* algorithm (Moore & Atkeson, 1993). As outlined in Section 2.2, prioritized sweeping works by maintaining an estimated transition model $\widehat{\mathrm{Pr}}$ and reward model $\widehat{R}$. Whenever an experience tuple $\langle s, a, r, t \rangle$ is sampled, the estimated model at state $s$ can change; a Bellman backup is performed at $s$ to incorporate the revised model and some (usually fixed) number of additional backups are performed at selected states. States are selected using a *priority* that estimates the potential change in their values based on the changes precipitated by earlier backups. Effectively, computational resources (backups) are focused on those states that can most "benefit" from those backups.

Incorporating our ideas into prioritized sweeping simply requires the following changes:

- With each transition $\langle s, a, t \rangle$ the observer takes, the estimated model $\widehat{\mathrm{Pr}}_o(s, a, t)$ is updated and an augmented backup is performed at state $s$. Augmented backups are then performed at a fixed number of states using the usual priority queue implementation.

- With each observed mentor transition $\langle s, t \rangle$, the estimated model $\widehat{\mathrm{Pr}}_m(s, t)$ is updated and an augmented backup is performed at $s$. Augmented backups are then performed at a fixed number of states using the usual priority queue implementation.

Keeping samples of mentor behavior implements model extraction. Augmented backups integrate this information into the observer's value function, and performing augmented backups at observed transitions (in addition to experienced transitions) incorporates our focusing mechanism. The observer is not forced to "follow" or otherwise mimic the actions of the mentor directly. But it does back up value information along the mentor's trajectory as if it had. Ultimately, the observer must move to those states to discover which actions are to be used; in the meantime, important value information is being propagated that can guide its exploration.

Implicit imitation does not alter the long run theoretical convergence properties of the underlying reinforcement learning algorithm. The implicit imitation framework is orthogonal to $\varepsilon$-greedy exploration, as it alters only the definition of the "greedy" action, not when the greedy action is taken. Given a theoretically appropriate decay factor, the $\varepsilon$-greedy strategy will thus ensure that the distributions for the action models at each state are sampled infinitely often in the limit and converge to their true values. Since the extracted model from the mentor corresponds to one of the observer's own actions, its effect on the value function calculations is no different than the effect of the observer's own sampled action models. The confidence mechanism ensures that the model with more samples will eventually come to dominate if it is, in fact, better. We can therefore be sure that the convergence properties of reinforcement learning with implicit imitation are identical to that of the underlying reinforcement learning algorithm.

The benefit of implicit imitation lies in the way in which the models extracted from the mentor allow the observer to calculate a lower bound on the value function and use this lower bound to choose its greedy actions to move the agent towards higher-valued regions of state space. The result is quicker convergence to optimal policies and better short-term practical performance with respect to accumulated discounted reward while learning.

### 4.2.7 EXTENSIONS

The implicit imitation model can easily be extended to extract model information from multiple mentors, mixing and matching pieces extracted from each mentor to achieve good results. It does this by searching, at each state, the set of mentors it knows about to find the mentor with the highest value estimate. The value estimate of the "best" mentor is then compared using the confidence test described above with the observer's own value estimate. The formal expression of the algorithm is given by the *multi-augmented Bellman equation*:

$$V(s) = R_o(s) + \gamma \max \left\{ \max_{a \in \mathcal{A}_o} \left\{ \sum_{t \in \mathcal{S}} \mathrm{Pr}_o(s, a, t) V(t) \right\}, \right.$$

$$\left. \max_{m \in \mathcal{M}} \sum_{t \in \mathcal{S}} \mathrm{Pr}_m(s, t) V(t) \right\} \quad (11)$$

where $\mathcal{M}$ is the set of candidate mentors. Ideally, confidence estimates should be taken into account when comparing mentor estimates with each other, as we may get a mentor with a high mean value estimate but large variance. If the observer has any experience with the state at all, this mentor will likely be rejected as having poorer quality information than the observer already has from its own experience. The observer might have been better off picking a mentor with a lower mean but more confident estimate that would have succeeded in the test against the observer's own model. In the interests of simplicity, however, we investigate multiple mentor combination without confidence testing.

Up to now, we have assumed no action costs (i.e., the agent's rewards depend only on the state and not on the action selected in the state); however, we can use more general reward functions (e.g., where reward has the form $R(s, a)$). The difficulty lies in backing up action costs when the mentor's chosen action is unknown. In Section 4.2.5 we defined the closest action function $\kappa$. The $\kappa$ function can be used to choose the appropriate reward. The augmented Bellman equation with generalized rewards takes the following form:

$$V(s) = \max \left\{ \max_{a \in \mathcal{A}_o} \left\{ R_o(s, a) + \gamma \sum_{t \in \mathcal{S}} \mathrm{Pr}_o(s, a, t) V(t) \right\}, \right.$$

$$R_o(s, \kappa(s)) + \gamma \sum_{t \in \mathcal{S}} \mathrm{Pr}_m(s, t) V(t) \Big\}$$

We note that Bayesian methods could be used could be used to estimate action costs in the mentor's chain as well. In any case, the generalized reward augmented equation can readily be amended to use confidence estimates in a similar fashion to the transition model.

### 4.3 Empirical Demonstrations

The following empirical tests incorporate model extraction and our focusing mechanism into prioritized sweeping. The results illustrate the types of problems and scenarios in which implicit imitation can provide advantages to a reinforcement learning agent. In each of the experiments, an expert mentor is introduced into the experiment to serve as a model for the observer. In each case, the mentor is following an $\varepsilon$-greedy policy with a very small $\varepsilon$ (on the order of 0.01). This tends to cause the mentor's trajectories to lie within a "cluster" surrounding optimal trajectories (and reflect good if not optimal policies). Even with a small amount of exploration and some environment stochasticity, mentors generally do not "cover" the entire state space, so confidence testing is important.

In all of these experiments, prioritized sweeping is used with a fixed number of backups per observed or experienced sample.[12] $\varepsilon$-greedy exploration is used with decaying $\varepsilon$. Observer agents are given uniform Dirichlet priors and Q-values are initialized to zero. Observer agents are compared to control agents that do not benefit from a mentor's experience, but are otherwise identical (implementing prioritized sweeping with similar parameters and exploration policies). The tests are all performed on stochastic grid world domains, since these make it clear to what extent the observer's and mentor's optimal policies overlap (or fail to). In Figure 2, a simple $10 \times 10$ example shows a start and end state on a grid. A typical optimal mentor trajectory is illustrated by the solid line between the start and end states. The dotted line shows that a typical mentor-influenced trajectory will be quite similar to the observed mentor trajectory. We assume eight-connectivity between cells so that any state in the grid has nine neighbors including itself, but agents have only four possible actions. In most experiments, the four actions move the agent in the compass directions (North, South, East and West), although the agent will not initially know which action does which. We focus primarily on whether imitation improves performance during learning, since the learner will converge to an optimal policy whether it uses imitation or not.

### 4.3.1 EXPERIMENT 1: THE IMITATION EFFECT

In our first experiment we compare the performance of an observer using model extraction and an expert mentor with the performance of a control agent using independent reinforcement learning. Given the uniform nature of this grid world and the lack of intermediate rewards, confidence testing is not required. Both agents attempt to learn a policy that maximizes discounted return in a $10 \times 10$ grid world. They start in the upper-left corner and seek a goal with value 1.0 in the lower-right corner. Upon reaching the goal, the agents are restarted in the upper-left corner. Generally the mentor will follow a similar if not identical trajectory each run, as the mentors were trained using a greedy strategy that leaves one path slightly more highly valued than the rest. Action dynamics are noisy, with the "intended" direction being realized 90% of the time, and one of the other directions taken otherwise (uniformly). The discount factor is 0.9. In Figure 3, we plot the cumulative number of goals obtained over the previous 1000 time steps for the observer "Obs" and control "Ctrl" agents (results are averaged over ten runs). The observer is able to quickly incorporate a policy learned from the mentor into its value estimates. This results in a steeper learning curve. In contrast, the control agent slowly explores the space to build a model first. The "Delta" curve shows the difference in performance between the agents. Both agents converge to the same optimal value function.

---

12. Generally, the number of backups was set to be roughly equal to the length of the optimal "noise-free" path.

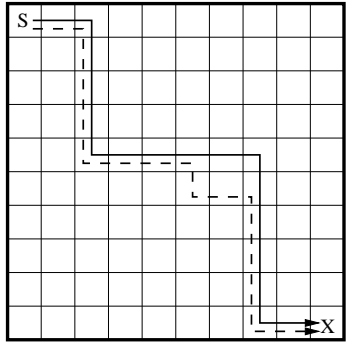

Figure 2: A simple grid world with start state $S$ and goal state $X$

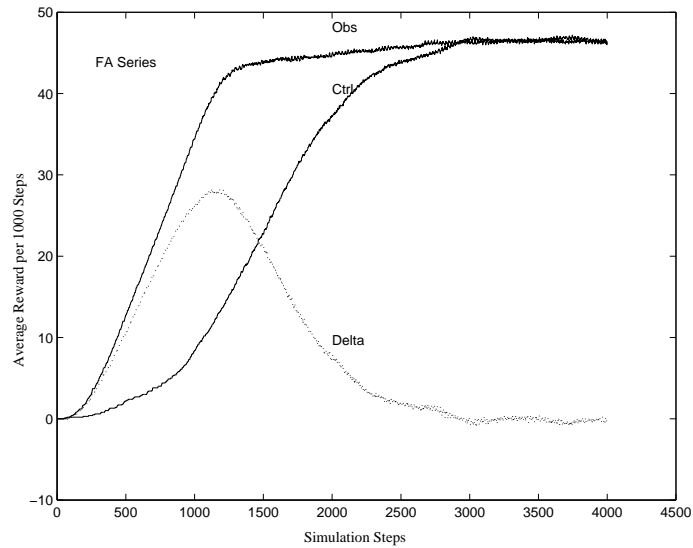

Figure 3: Basic observer and control agent comparisons

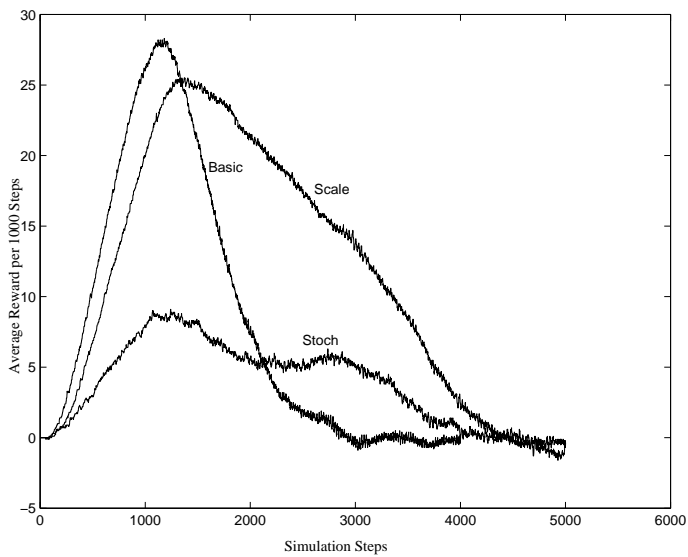

Figure 4: Delta curves showing the influence of domain size and noise

### 4.3.2 EXPERIMENT 2: SCALING AND NOISE

The next experiment illustrates the sensitivity of imitation to the size of the state space and action noise level. Again, the observer uses model-extraction but not confidence testing. In Figure 4, we plot the Delta curves (i.e., *difference* in performance between observer and control agents) for the "Basic" scenario just described, the "Scale" scenario in which the state space size is increased 69 percent (to a $13 \times 13$ grid), and the "Stoch" scenario in which the noise level is increased to 40 percent (results are averaged over ten runs). The total gain represented by the area under the curves for the observer and the non-imitating prioritized sweeping agent increases with the state space size. This reflects Whitehead's (1991a) observation that for grid worlds, exploration requirements can increase quickly with state space size, but that the optimal path length increases only linearly. Here we see that the guidance of the mentor can help more in larger state spaces.

Increasing the noise level reduces the observer's ability to act upon the information received from the mentor and therefore erodes its advantage over the control agent. We note, however, that the benefit of imitation degrades gracefully with increased noise and is present even at this relatively extreme noise level.

### 4.3.3 EXPERIMENT 3: CONFIDENCE TESTING

Sometimes the observer's prior beliefs about the transition probabilities of the mentor can mislead the observer and cause it to generate inappropriate values. The confidence mechanism proposed in the previous section can prevent the observer from being fooled by misleading priors over the mentor's transition probabilities. To demonstrate the role of the confidence mechanism in implicit imitation, we designed an experiment based on the scenario illustrated in Figure 5. Again, the agent's task is to navigate from the top-left corner to the bottom-right corner of a $10 \times 10$ grid in order to attain a reward of $+1$. We have cre-ated a pathological scenario in which islands of high reward (+5) are enclosed by obstacles. Since the observer's priors reflect eight-connectivity and are uniform, the high-valued cells in the middle of each island are believed to be reachable from the states diagonally adjacent with some small prior probability. In reality, however, the agent's action set precludes this and the agent will therefore never be able to realize this value. The four islands in this scenario thus create a fairly large region in the center of the space with a high estimated value, which could potentially trap an observer if it persisted in its prior beliefs.

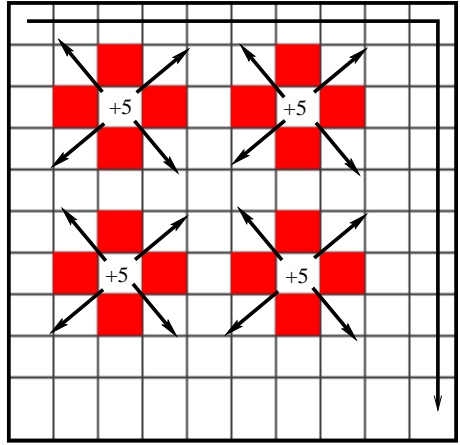

Figure 5: An environment with misleading priors

Notice that a standard reinforcement learner will "quickly" learn that none of its actions take it to the rewarding islands; in contrast, an implicit imitator using augmented backups could be fooled by its prior mentor model. If the mentor does not visit the states neighboring the island, the observer will not have any evidence upon which to change its prior belief that the mentor actions are equally likely to take one in any of the eight possible directions. The imitator may falsely conclude on the basis of the mentor action model that an action does exist which would allow it to access the islands of value. The observer therefore needs a confidence mechanism to detect when the mentor model is less reliable than its own model.

To test the confidence mechanism, we have the mentor follows a path around the outside of the obstacles so that its path cannot lead the observer out of the trap (i.e., it provides no evidence to the observer that the diagonal moves into the islands are not feasible). The combination of a high initial exploration rate and the ability of prioritized sweeping to spread value across large distances then virtually guarantees that the observer will be led to the trap. Given this scenario, we ran two observer agents and a control. The first observer used a confidence interval with width given by $5\sigma$, which, according to the Chebychev rule, should cover approximately 96 percent of an arbitrary distribution. The second observer was given a $0\sigma$ interval, which effectively disables confidence testing. The observer with no confidence testing consistently became stuck. Examination of the value function revealed consistent peaks within the trap region, and inspection of the agent state trajectories showed that it was stuck in the trap. The observer with confidence testing consistently escaped the trap. Observation of its value function over time shows that the trap formed, but faded away as the observer gained enough experience to with its own actions to allow it to ignore overcome erroneous priors over the mentor actions. In Figure 6, the performance of the observer with confidence testing is shown with the performance of the control agent (results are averaged over 10 runs). We see that the observer's performance is only slightly degraded from that of the unaugmented control agent even in this pathological case.

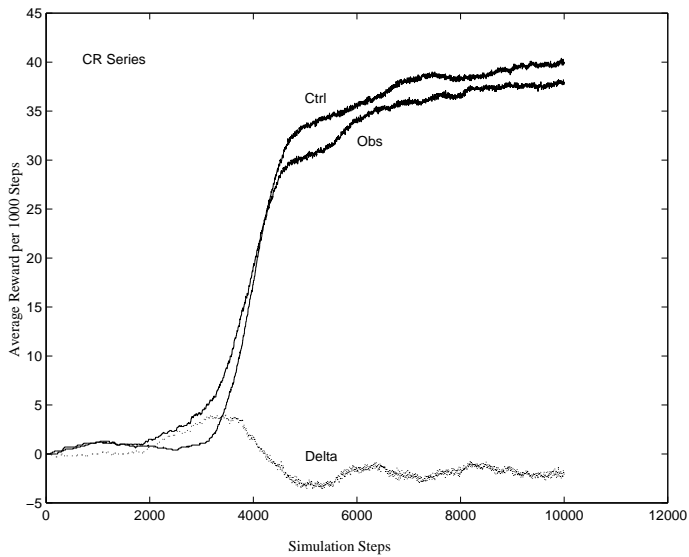

Figure 6: Misleading priors may degrade performance

### 4.3.4 EXPERIMENT 4: QUALITATIVE DIFFICULTY

The next experiment demonstrates how the potential gains of imitation can increase with the (qualitative) difficulty of the problem. The observer employs both model extraction and confidence testing, though confidence testing will not play a significant role here.[13] In the "maze" scenario, we introduce obstacles in order to increase the difficulty of the learning problem. The maze is set on a $25 \times 25$ grid (Figure 7) with 286 obstacles complicating the agent's journey from the top-left to the bottom-right corner. The optimal solution takes the form of a snaking 133-step path, with distracting paths (up to length 22) branching off from the solution path necessitating frequent backtracking. The discount factor is 0.98. With 10 percent noise, the optimal goal-attainment rate is about six goals per 1000 steps.

From the graph in Figure 8 (with results averaged over ten runs), we see that the control agent takes on the order of 200,000 steps to build a decent value function that reliably leads to the goal. At this point, it is only achieving four goals per 1000 steps on average, as its exploration rate is still reasonably high (unfortunately, decreasing exploration more quickly leads to slower value function formation). The imitation agent is able to take advantage of the mentor's expertise to build a reliable value function in about 20,000 steps. Since the control agent has been unable to reach the goal at all in the first 20,000 steps, the Delta between the control and the imitator is simply equal to the imitator's performance. The imitator can quickly achieve the optimal goal attainment rate of six goals per 1000 steps, as its exploration rate decays much more quickly.

---

13. The mentor does not provide evidence about some path choices in this problem, but there are no intermediate rewards which would cause the observer to make use of the misleading mentor priors at these states.

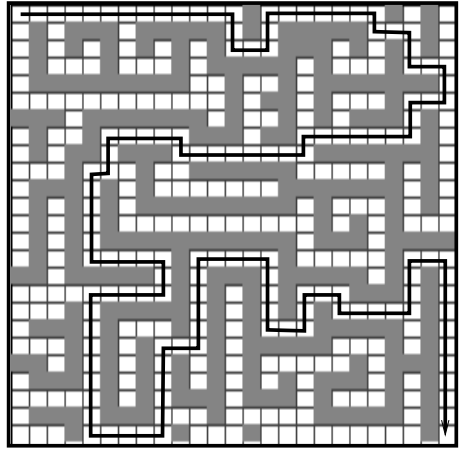

Figure 7: A complex maze

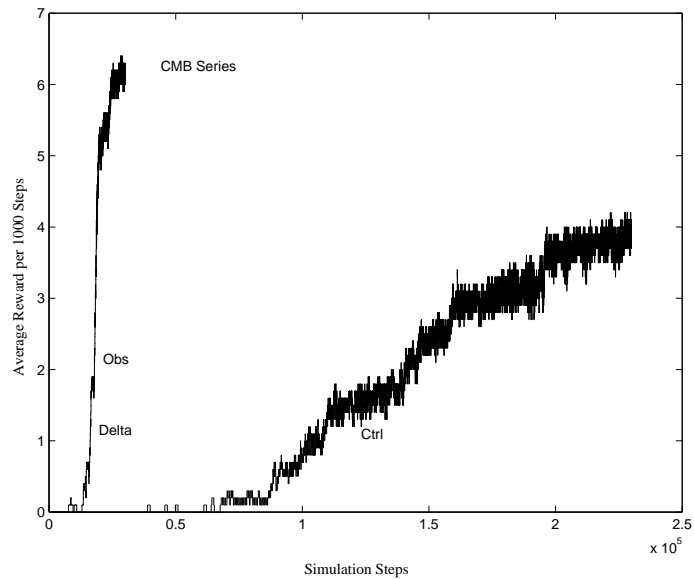

Figure 8: Imitation in a complex space

### 4.3.5 EXPERIMENT 5: IMPROVING SUBOPTIMAL POLICIES BY IMITATION

The augmented backup rule does not require that the reward structure of the mentor and observer be identical. There are many useful scenarios where rewards are dissimilar but the value functions and policies induced share some structure. In this experiment, we demonstrate one interesting scenario in which it is relatively easy to find a suboptimal solution, but difficult to find the optimal solution. Once the observer finds this suboptimal path, however, it is able to exploit its observations of the mentor to see that there is a shortcut that significantly shortens the path to the goal. The structure of the scenario is shown in Figure 9. The suboptimal solution lies on the path from location 1 around the "scenic route" to location 2 and on to the goal at location 3. The mentor takes the vertical path from location 4 to location 5 through the shortcut.[14] To discourage the use of the shortcut by novice agents, it is lined with cells (marked "*") from which the agent immediately jumps back to the start state. It is therefore difficult for a novice agent executing random exploratory moves to make it all the way to the end of the shortcut and obtain the value which would reinforce its future use. Both the observer and control therefore generally find the scenic route first.

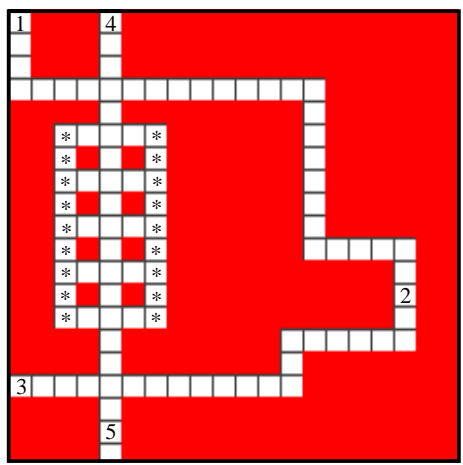

Figure 9: A maze with a perilous shortcut

In Figure 10, the performance (measured using goals reached over the previous 1000 steps) of the control and observer are compared (averaged over ten runs), indicating the value of these observations. We see that the observer and control agent both find the longer scenic route, though the control agent takes longer to find it. The observer goes on to find the shortcut and increases its return to almost double the goal rate. This experiment shows that mentors can improve observer policies even when the observer's goals are not on the mentor's path.

### 4.3.6 EXPERIMENT 6: MULTIPLE MENTORS

The final experiment illustrates how model extraction can be readily extended so that the observer can extract models from multiple mentors and exploit the most valuable parts of each. Again, the observer employs model extraction and confidence testing. In Figure 11, the learner must move from start location 1 to goal location 4. Two expert agents with different start and goal states serve as potential mentors. One mentor repeatedly moves from location 3 to location 5 along the dotted line, while a second mentor departs from location 2 and ends at location 4 along the dashed line. In this experiment, the observer must combine the information from the examples provided by the two mentors with independent exploration of its own in order to solve the problem.

---

14. A mentor proceeding from 5 to 4 would not provide guidance without prior knowledge that actions are reversible.

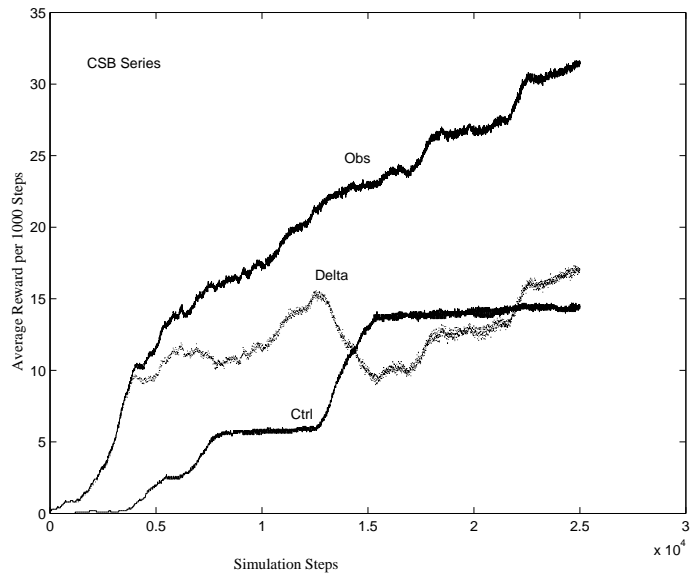

Figure 10: Transfer with non-identical rewards

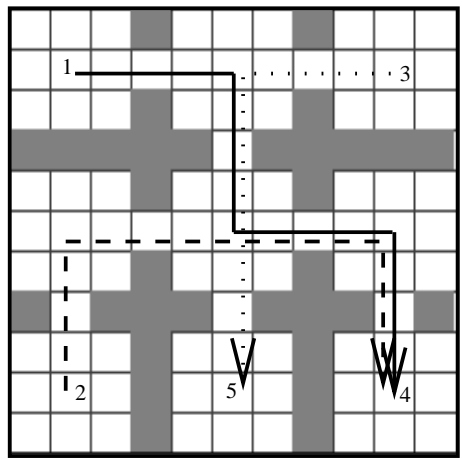

Figure 11: Multiple mentors scenario

In Figure 12, we see that the observer successfully pulls together these information sources in order to learn much more quickly than the control agent (results are averaged over 10 runs). We see that the use of a value-based technique allows the observer to choose which mentor's influence to use on a state-by-state basis in order to get the best solution to the problem.

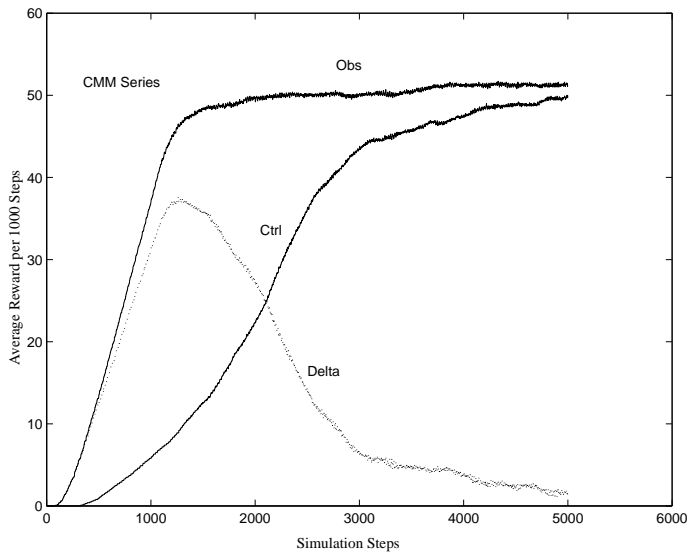

Figure 12: Learning from multiple mentors

# 5. Implicit Imitation in Heterogeneous Settings

When the homogeneity assumption is violated, the implicit imitation framework described above can cause the learner's convergence rate to slow dramatically and, in some cases, cause the learner to become stuck in a small neighborhood of state space. In particular, if the learner is unable to make the same state transition (or a transition with the same probability) as the mentor at a given state, it may drastically overestimate the value of that state. The inflated value estimate causes the learner to return repeatedly to this state even though its exploration will never produce a feasible action that attains the inflated estimated value. There is no mechanism for removing the influence of the mentor's Markov chain on value estimates—the observer can be extremely (and correctly) confident in its observations about the mentor's model. The problem lies in the fact that the augmented Bellman backup is justified by the assumption that the observer can duplicate *every* mentor action. That is, at each state $s$, there is some $a \in \mathcal{A}$ such that $\Pr_o(s, a, t) = \Pr_m(s, t)$ for all $t$. When an equivalent action $a$ does not exist, there is no guarantee that the value calculated using the mentor action model can, in fact, be achieved.

## 5.1 Feasibility Testing

In such heterogeneous settings, we can prevent "lock-up" and poor convergence through the use of an explicit *action feasibility test*: before an augmented backup is performed at $s$, the observer tests whether the mentor's action $a_m$ "differs" from each of its actions at $s$, given its current estimated models. If so, the augmented backup is suppressed and a standard Bellman backup is used to update the value function.[15] By default, mentor actions are assumed to be feasible for the observer; however, once the observer is reasonably confident that $a_m$ is infeasible at state $s$, augmented backups are suppressed at $s$.

---

15. The decision is binary; but we could envision a smoother decision criterion that measures the extent to which the mentor's action can be duplicated.

Recall that uncertainty about the agent's true transition probabilities are captured by a Dirichlet distribution derived from sampled transitions. Comparing $a_m$ with $a_o$ is effected by a difference of means test with respect to the corresponding Dirichlets. This is complicated by the fact that Dirichlets are highly non-normal for small parameter values and transition distributions are multinomial. We deal with the non-normality by requiring a minimum number of samples and using robust Chebychev bounds on the pooled variance of the distributions to be compared. Conceptually, we will evaluate Equation 12:

$$\frac{|\Pr_o(s,a_o,t) - \Pr_m(s,t)|}{\sqrt{\frac{n_o(s,a_o,t)\sigma^2_{omodel}(s,a_o,t) + n_m(s,t)\sigma^2_{mmodel}(s,t)}{n_o(s,a_o,t) + n_m(s,t)}}} > Z_{\alpha/2} \tag{12}$$

Here $Z_{\alpha/2}$ is the critical value of the test. The parameter $\alpha$ is the significance of the test, or the probability that we will falsely reject two actions as being different when they are actually the same. Given our highly non-normal distributions early in the training process, the appropriate $Z$ value for a given $\alpha$ can be computed from Chebychev's bound by solving $2\alpha = 1 - \frac{1}{Z^2}$ for $Z_{\alpha/2}$.

When we have too few samples to do an accurate test, we persist with augmented backups (embodying our default assumption of homogeneity). If the value estimate is inflated by these backups, the agent will be biased to obtain additional samples, which will then allow the agent to perform the required feasibility test. Our assumption is therefore self-correcting. We deal with the multivariate complications by performing the *Bonferroni test* (Seber, 1984), which has been shown to give good results in practice (Mi & Sampson, 1993), is efficient to compute, and is known to be robust to dependence between variables. A Bonferroni hypothesis test is obtained by conjoining several single variable tests. Suppose the actions $a_o$ and $a_m$ result in $r$ possible successor states, $s_1, \cdots, s_r$ (i.e., $r$ transition probabilities to compare). For each $s_i$, the hypothesis $E_i$ denotes that $a_o$ and $a_m$ have the same transition probability to successor state $s_i$; that is $\Pr(s, a_m, s_i) = \Pr(s, a_o, s_i)$. We let $\bar{E}_i$ denote the complementary hypothesis (i.e., that the transition probabilities differ). The Bonferroni inequality states:

$$\Pr\left[\bigcap_{i=1}^{r} E_i\right] \geq 1 - \sum_{i=1}^{r} \Pr\left[\bar{E}_i\right]$$

Thus we can test the joint hypothesis $\bigcap_{i=1}^{r} E_i$—the two action models are the same—by testing each of the $r$ complementary hypotheses $\bar{E}_i$ at confidence level $\alpha/r$. If we reject any of the hypotheses we reject the notion that the two actions are equal with confidence at least $\alpha$. The mentor action $a_m$ is deemed infeasible if for every observer action $a_o$, the multivariate Bonferroni test just described rejects the hypothesis that the action is the same as the mentor's.

Pseudo-code for the Bonferroni component of the feasibility test appears in Table 2. It assumes a sufficient number of samples. For efficiency reasons, we cache the results of the feasibility testing. When the duplication of the mentor's action at state $s$ is first determined to be infeasible, we set a flag for state $s$ to this effect.

```
FUNCTION feasible(m,s) : Boolean
    FOR each a in A_o do
        allSuccessorProbsSimilar = true
        FOR each t in successors(s) do
            μ_Δ = |Pr_o(s,a,t) − Pr_m(s,t)|
            z_Δ = μ_Δ / √( (n_o(s,a,t)∗σ²_omodel(s,a,t) + n_m(s,t)σ²_mmodel(s,t)) / (n_o(s,a,t) + n_m(s,t)) )
            IF z_Δ > z_{α/2r}
                allSuccessorProbsSimilar = false
        END FOR
        IF allSuccessorProbsSimilar
            return true
    END FOR
    RETURN false
```

Table 2: Action Feasibility Testing

## 5.2 $k$-step Similarity and Repair

Action feasibility testing essentially makes a strict decision as to whether the agent can duplicate the mentor's action at a specific state: once it is decided that the mentor's action is infeasible, augmented backups are suppressed and all potential guidance offered is eliminated at that state. Unfortunately, the strictness of the test results in a somewhat impoverished notion of similarity between mentor and observer. This, in turn, unnecessarily limits the transfer between mentor and observer. We propose a mechanism whereby the mentor's influence may persist even if the specific action it chooses is not feasible for the mentor; we instead rely on the possibility that the observer may *approximately* duplicate the mentor's trajectory instead of exactly duplicating it.

Suppose an observer has previously constructed an estimated value function using augmented backups. Using the mentor action model (i.e., the mentor's chain $\mathrm{Pr}_m(s,t)$), a high value has been calculated for state $s$. Subsequently, suppose the mentor's action at state $s$ is judged to be infeasible. This is illustrated in Figure 13, where the estimated value at state $s$ is originally due to the mentor's action $\pi_m(s)$, which for the sake of illustration moves with high probability to state $t$, which itself can lead to some highly-rewarding region of state space. After some number of experiences at state $s$, however, the learner concludes that the action $\pi_m(s)$—and the associated high probability transition to $t$—is not feasible.

At this point, one of two things must occur: either (a) the value calculated for state $s$ and its predecessors will "collapse" and all exploration towards highly-valued regions beyond state $s$ ceases; or (b) the estimated value drops slightly but exploration continues towards the highly-valued regions. The latter case may arise as follows. If the observer has previously explored in the vicinity of state $s$, the observer's own action model may be sufficiently developed that they still connect the higher value-regions beyond state $s$ to state $s$ through Bellman backups. For example, if the learner has sufficient experience to have learned that the highly-valued region can be reached through the alternative trajectory $s - u - v - w$, the newly discovered infeasibility of the mentor's transition $s - t$ will not have a deleterious effect on the value estimate at $s$. If $s$ is highly-valued, it is likely that states close to the mentor's trajectory will be explored to some degree. In this case, state $s$ will not be as highly-valued as it was when using the mentor's action model, but it will still be valued highly enough that it will likely to guide further exploration toward the area. We call this alternative (in this case $s - u - v - w$) to the mentor's action a *bridge*, because it allows value from higher value regions to "flow over" an infeasible mentor transition. Because the bridge was formed without the intention of the agent, we call this process *spontaneous bridging*.

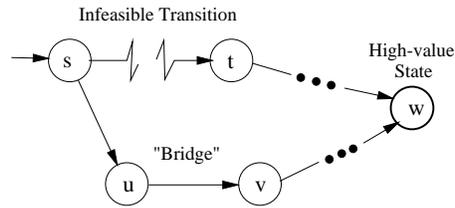

Figure 13: An alternative path can bridge value backups around infeasible paths

Where a spontaneous bridge does not exist, the observer's own action models are generally undeveloped (e.g., they are close to their uniform prior distributions). Typically, these undeveloped models assign a small probability to every possible outcome and therefore diffuse value from higher valued regions and lead to a very poor value estimate for state $s$. The result is often a dramatic drop in the value of state $s$ and all of its predecessors; and exploration towards the highly-valued region through the neighborhood of state $s$ ceases. In our example, this could occur if the observer's transition models at state $s$ assign low probability (e.g., close to prior probability) of moving to state $u$ due to lack of experience (or similarly if the surrounding states, such as $u$ or $v$, have been insufficiently explored).

The spontaneous bridging effect motivates a broader notion of similarity. When the observer can find a "short" sequence of actions that bridges an infeasible action on the mentor's trajectory, the mentor's example can still provide extremely useful guidance. For the moment, we assume a short path is any path of length no greater than some given integer $k$. We say an observer is *k-step similar to a mentor* at state $s$ if the observer can duplicate in $k$ or fewer steps the mentor's nominal transition at state $s$ with "sufficiently high" probability.

Given this notion of similarity, an observer can now test whether a spontaneous bridge exists and determine whether the observer is in danger of value function collapse and the concomitant loss of guidance if it decides to suppress an augmented backup at state $s$. To do this, the observer initiates a reachability analysis starting from state $s$ using its own action model $\mathrm{Pr}_o(s, a, t)$ to determine if there is a sequence of actions with leads with sufficiently high probability from state $s$ to some state $t$ on the mentor's trajectory downstream of the infeasible action.[16] If a $k$-step bridge already exists, augmented backups can be safely suppressed at state $s$. For efficiency, we maintain a flag at each state to mark it as "bridged." Once a state is known to be bridged, the $k$-step reachability analysis need not be repeated.

If a spontaneous bridge cannot be found, it might still be possible to intentionally set out to build one. To build a bridge, the observer must explore from state $s$ up to $k$-steps away, hoping to make contact with the mentor's trajectory downstream of the infeasible mentor action. We implement a single search attempt as a $k^2$-step random walk, which will result in a trajectory on average $k$ steps away from $s$ as long ergodicity and local connectivity assumptions are satisfied. In order for the search to occur, we must motivate the observer to return to the state $s$ and engage in repeated exploration. We could provide motivation to the observer by asking the observer to assume that the infeasible action will be repairable. The observer will therefore continue the augmented backups which support high-value estimates at the state $s$ and the observer will repeatedly engage in exploration from this point. The danger, of course, is that there may not in fact be a bridge, in which case the observer will repeat this search for a bridge indefinitely. We therefore need a mechanism to terminate the repair process when a $k$-step repair is infeasible. We could attempt to explicitly keep track of all of the possible paths open to the observer and all of the paths explicitly tried by the observer and determine the repair possibilities had been exhausted. Instead, we elect to follow a probabilistic search that eliminates the need for bookkeeping: if a bridge cannot be constructed within $n$ attempts of $k$-step random walk, the "repairability assumption" is judged falsified, the augmented backup at state $s$ is suppressed and the observer's bias to explore the vicinity of state $s$ is eliminated. If no bridge is found for state $s$, a flag is used to mark the state as "irreparable."

---

16. In a more general state space where ergodicity is lacking, the agent must consider predecessors of state $s$ up to $k$ steps before $s$ to guarantee that all $k$-step paths are checked.

This approach is, of course, a very naïve heuristic strategy; but it illustrates the basic import of bridging. More systematic strategies could be used, involving explicit "planning" to find a bridge using, say, local search (Alissandrakis, Nehaniv, & Dautenhahn, 2000). Another aspect of this problem that we do not address is the persistence of search for bridges. In a specific domain, after some number of unsuccessful attempts to find bridges, a learner may conclude that it is unable to reconstruct a mentor's behavior, in which case the search for bridges may be abandoned. This involves simple, higher-level inference, and some notion of (or prior beliefs about) "similarity" of capabilities. These notions could also be used to automatically determine parameter settings (discussed below).

The parameters $k$ and $n$ must be tuned empirically, but can be estimated given knowledge of the connectivity of the domain and prior beliefs about how similar (in terms of length of average repair) the trajectories of the mentor and observer will be. For instance, $n > 8k - 4$ seems suitable in an 8-connected grid world with low noise, based on the number of trajectories required to cover the perimeter states of a $k$-step rectangle around a state. We note that very large values of $n$ can reduce performance below that of non-imitating agents as it results in temporary "lock up."

Feasibility and $k$-step repair are easily integrated into the homogeneous implicit imitation framework. Essentially, we simply elaborate the conditions under which the augmented backup will be employed. Of course, some additional representation will be introduced to keep track of whether a state is feasible, bridged, or repairable, and how many repair attempts have been made. The action selection mechanism will also be overridden by the bridge-building algorithm when required in order to search for a bridge. Bridge building always terminates after $n$ attempts, however, so it cannot affect long run convergence. All other aspects of the algorithm, however, such as the exploration policy, are unchanged.

The complete elaborated decision procedure used to determine when augmented backups will be employed at state $s$ with respect to mentor $m$ appears in Table 3. It uses some internal state to make its decisions. As in the original model, we first check to see if the observer's experience-based calculation for the value of the state supersedes the mentor-

```
FUNCTION use_augmented?(s,m) : Boolean
    IF V_o(s) ≻ V_m(s) THEN RETURN false
    ELSE IF feasible(s, m) THEN RETURN true
    ELSE IF bridged(s, m) THEN RETURN false
    ELSE IF reachable(s, m) THEN
        bridged(s,m) := true
        RETURN false
    ELSE IF not repairable(s, m) THEN return false
    ELSE % we are searching
        IF 0 < search_steps(s, m) < k THEN % search in progress
            return true
        IF search_steps(s, m) > k THEN % search failed
            IF attempts(s) > n THEN
                repairable(s) = false
                RETURN false
            ELSE
                reset_search(s,m)
                attempts(s) := attempts(s) + 1
                RETURN true

        attempts(s) :=1 % initiate first attempt of a search
        initiate-search(s)
        RETURN true
```

Table 3: Elaborated augmented backup test

based calculation; if so, then the observer uses its own experience-based calculation. If the mentor's action is feasible, then we accept the value calculated using the observation-based value function. If the action is infeasible we check to see if the state is bridged. The first time the test is requested, a reachability analysis is performed, but the results will be drawn from a cache for subsequent requests. If the state has been bridged, we suppress augmented backups, confident that this will not cause value function collapse. If the state is not bridged, we ask if it is repairable. For the first $n$ requests, the agent will attempt a $k$-step repair. If the repair succeeds, the state is marked as bridged. If we cannot repair the infeasible transition, we mark it not-repairable and suppress augmented backups.

We may wish to employ implicit imitation with feasibility testing in a multiple-mentor scenario. The key change from implicit imitation without feasibility testing is that the observer will only imitate feasible actions. When the observer searches through the set of mentors for the one with the action that results in the highest value estimate, the observer must consider only those mentors whose actions are still considered feasible (or assumed to be repairable).

## 5.3 Empirical Demonstrations

In this section, we empirically demonstrate the utility of feasibility testing and $k$-step repair and show how the techniques can be used to surmount both differences in actions between agents and small local differences in state-space topology. The problems here have been chosen specifically to demonstrate the necessity and utility of both feasibility testing and $k$-step repair.

### 5.3.1 Experiment 1: Necessity of Feasibility Testing

Our first experiment shows the importance of feasibility testing in implicit imitation when agents have heterogeneous actions. In this scenario, all agents must navigate across an obstacle-free, $10 \times 10$ grid world from the upper-left corner to a goal location in the lower-right. The agent is then reset to the upper-left corner. The first agent is a mentor with the "NEWS" action set (North, South, East, and West movement actions). The mentor is given an optimal stationary policy for this problem. We study the performance of three learners, each with the "Skew" action set (N, S, NE, SW) and unable to duplicate the mentor exactly (e.g., duplicating a mentor's E-move requires the learner to move NE followed by S, or move SE then N). Due to the nature of the grid world, the control and imitation agents will actually have to execute more actions to get to the goal than the mentor and the optimal goal rate for both the control and imitator are therefore lower than that of the mentor. The first learner employs implicit imitation *with* feasibility testing, the second uses imitation *without* feasibility testing, and the third control agent uses no imitation (i.e., is a standard reinforcement learning agent). All agents experience limited stochasticity in the form of a 5% chance that their action will be randomly perturbed. As in the last section, the agents use model-based reinforcement learning with prioritized sweeping. We set $k = 3$ and $n = 20$.

The effectiveness of feasibility testing in implicit imitation can be seen in Figure 14. The horizontal axis represents time in simulation steps and the vertical axis represents the average number of goals achieved per 1000 time steps (averaged over 10 runs). We see that the imitation agent with feasibility testing converges much more quickly to the optimal goal-attainment rate than the other agents. The agent without feasibility testing achieves sporadic success early on, but frequently "locks up" due to repeated attempts to duplicate infeasible mentor actions. The agent still manages to reach the goal from time to time, as the stochastic actions do not permit the agent to become permanently stuck in this obstacle-free scenario. The control agent without any form of imitation demonstrates a significant delay in convergence relative to the imitation agents due to the lack of any form of guidance, but easily surpasses the agent without feasibility testing in the long run. The more gradual slope of the control agent is due to the higher variance in the control agent's discovery time for the optimal path, but both the feasibility-testing imitator and the control agent converge to optimal solutions. As shown by the comparison of the two imitation agents, feasibility testing is necessary to adapt implicit imitation to contexts involving heterogeneous actions.

### 5.3.2 Experiment 2: Changes to State Space

We developed feasibility testing and bridging primarily to deal with the problem of adapting to agents with heterogeneous actions. The same techniques, however, can be applied to agents with differences in their state-space connectivity (ultimately, these are equivalent notions). To test this, we constructed a domain where all agents have the *same* NEWS action set, but we alter the environment of the learners by introducing obstacles that aren't present for the mentor. In Figure 15, the learners find that the mentor's path is obstructed by obstacles. Movement toward an obstacle causes a learner to remain in its current state. In this sense, its action has a different effect than the mentor's.

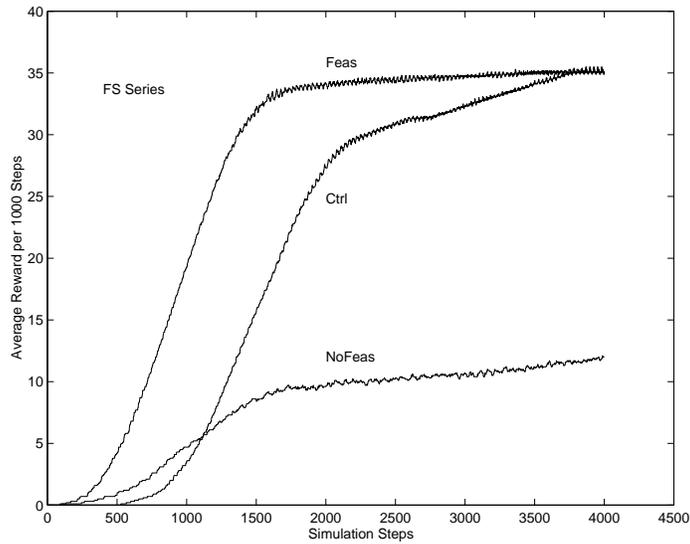

Figure 14: Utility of feasibility testing

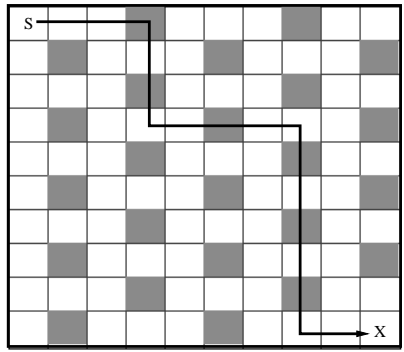

Figure 15: Obstacle map and mentor path

In Figure 16, we see that the results are qualitatively similar to the previous experiment. In contrast to the previous experiment, both imitator and control use the "NEWS" action set and therefore have a shortest path with the same length as that of the mentor. Consequently, the optimal goal rate of the imitators and control is higher than in the previous experiment. The observer without feasibility testing has difficulty with the maze, as the value function augmented by mentor observations consistently leads the observer to states whose path to the goal is directly blocked. The agent with feasibility testing quickly discovers that the mentor's influence is inappropriate at such states. We conclude that local differences in state are well handled by feasibility testing.

Next, we demonstrate how feasibility testing can completely generalize the mentor's trajectory. Here, the mentor follows a path which is completely infeasible for the imitating agent. We fix the mentor's path for all runs and give the imitating agent the maze shown in Figure 17 in which all but two of the states the mentor visits are blocked by an obstacle. The imitating agent is able to use the mentor's trajectory for guidance and builds its own parallel trajectory which is completely disjoint from the mentor's.

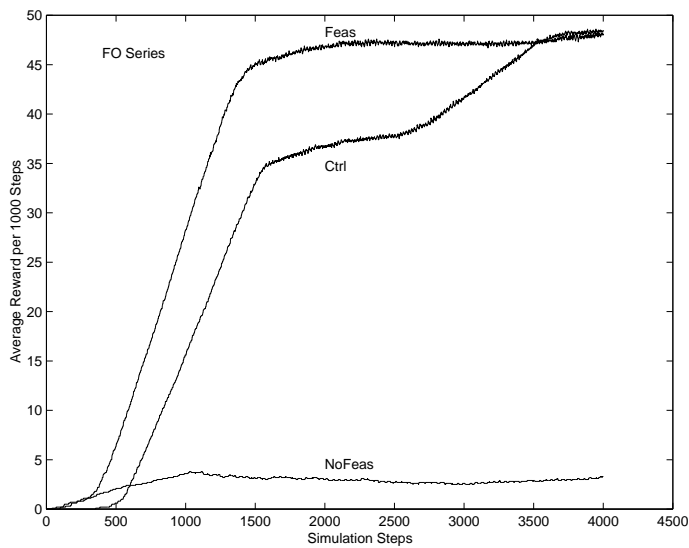

Figure 16: Interpolating around obstacles

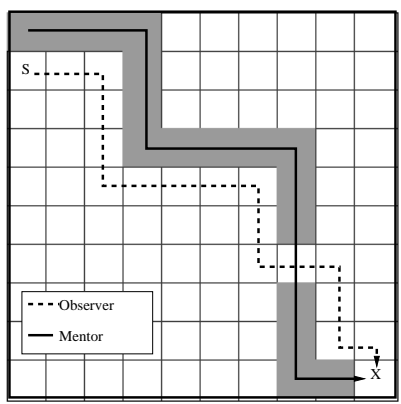

Figure 17: Parallel generalization

The results in Figure 18 show that gain of the imitator with feasibility testing over the control agent diminishes, but still exists marginally when the imitator is forced to generalize a completely infeasible mentor trajectory. The agent without feasibility testing does very poorly, even when compared to the control agent. This is because it gets stuck around the doorway. The high value gradient backed up along the mentor's path becomes accessible to the agents at the doorway. The imitation agent with feasibility will conclude that it cannot proceed south from the doorway (into the wall) and it will then try a different strategy. The imitator without feasibility testing never explores far enough away from the doorway to setup an independent value gradient that will guide it to the goal. With a slower decay schedule for exploration, the imitator without feasibility testing would find the goal, but this would still reduce its performance below that of the imitator with feasibility testing. The imitator with feasibility testing makes use of its prior beliefs that it can follow the mentor to backup value perpendicular to the mentor's path. A value gradient will therefore form parallel to the infeasible mentor path and the imitator can follow along side the infeasible path towards the doorway where it makes the necessary feasibility test and then proceeds to the goal.

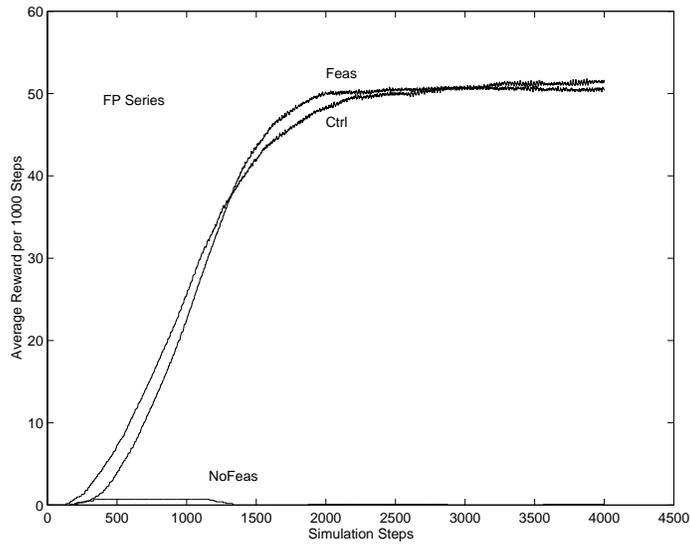

Figure 18: Parallel generalization results

As explained earlier, in simple problems there is a good chance that the informal effects of prior value leakage and stochastic exploration may form bridges before feasibility testing cuts off the value propagation that guides exploration. In more difficult problems where the agent spends a lot more time exploring, it will accumulate sufficient samples to conclude that the mentor's actions are infeasible long before the agent has constructed its own bridge. The imitator's performance would then drop down to that of an unaugmented reinforcement learner.

To demonstrate bridging, we devised a domain in which agents must navigate from the upper-left corner to the bottom-right corner, across a "river" which is three steps wide and exacts a penalty of −0.2 per step (see Figure 19). The goal state is worth 1.0. In the figure, the path of the mentor is shown starting from the top corner, proceeding along the edge of the river and then crossing the river to the goal. The mentor employs the "NEWS" action set. The observer uses the "Skew" action set (N, NE, S, SW) and attempts to reproduce the mentor trajectory. It will fail to reproduce the critical transition at the border of the river (because the "East" action is infeasible for a "Skew" agent). The mentor action can no longer be used to backup value from the rewarding state and there will be no alternative paths because the river blocks greedy exploration in this region. Without bridging or an optimistic and lengthly exploration phase, observer agents quickly discover the negative states of the river and curtail exploration in this direction before actually making it across.

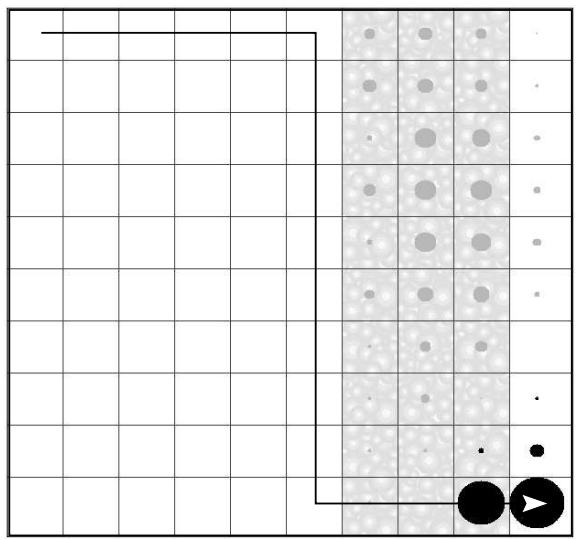

Figure 19: River scenario

If we examine the value function estimate (after 1000 steps) of an imitator with feasibility testing but no repair capabilities, we see that, due to suppression by feasibility testing, the darkly shaded high-value states in Figure 19 (backed up from the goal) terminate abruptly at an infeasible transition without making it across the river. In fact, they are dominated by the lighter grey circles showing negative values. In this experiment, we show that bridging can prolong the exploration phase in just the right way. We employ the $k$-step repair procedure with $k = 3$.

Examining the graph in Figure 20, we see that both imitation agents experience an early negative dip as they are guided deep into the river by the mentor's influence. The agent without repair eventually decides the mentor's action is infeasible, and thereafter avoids the river (and the possibility of finding the goal). The imitator with repair also discovers the mentor's action to be infeasible, but does not immediately dispense with the mentor's guidance. It keeps exploring in the area of the mentor's trajectory using a random walk, all the while accumulating a negative reward until it suddenly finds a bridge and rapidly converges on the optimal solution.[17] The control agent discovers the goal only once in the ten runs.

## 6. Applicability

The simple experiments presented above demonstrate the major qualitative issues confronting an implicit imitation agent and how the specific mechanisms of implicit imitation address these issues. In this section, we examine how the assumptions and the mechanisms we presented in the previous sections determine the types of problems suitable for implicit imitation. We then present several dimensions that prove useful for predicting the performance of implicit imitation in these types of problems.

---

17. While repair steps take place in an area of negative reward in this scenario, this need not be the case. Repair doesn't *imply* short-term negative return.

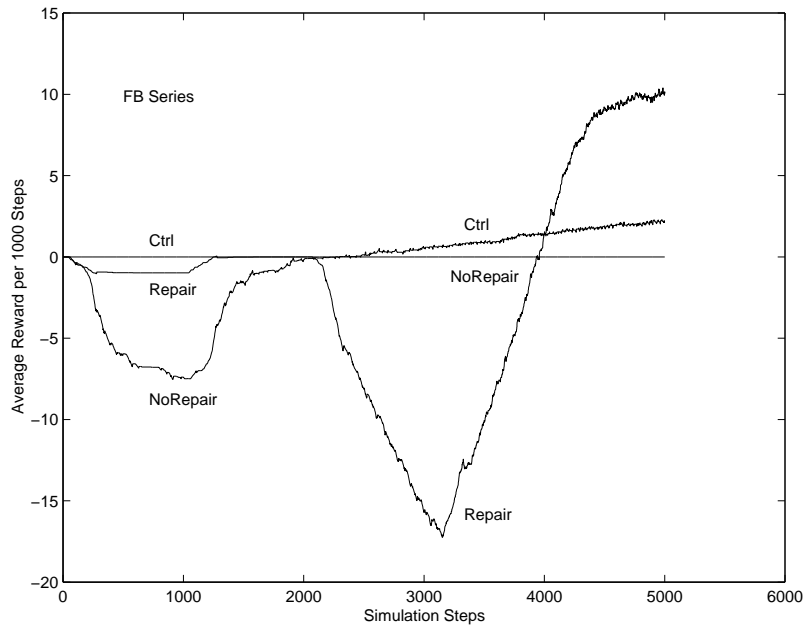

Figure 20: Utility of bridging

We have already identified a number of assumptions under which implicit imitation is applicable—some assumptions under which other models of imitation or teaching cannot be applied, and some assumptions that restrict the applicability of our model. These include: lack of explicit communication between mentors and observer; independent objectives for mentors and observer; full observability of mentors by observer; unobservability of mentors' actions; and (bounded) heterogeneity. Assumptions such as full observability are necessary for our model—as formulated—to work (though we discuss extension to the partially observable case in Section 7). Assumptions of lack of communication and unobservable actions extend the applicability of implicit imitation beyond other models in the literature; if these conditions do not hold, a simpler form of explicit communication may be preferable. Finally, the assumptions of bounded heterogeneity and independent objectives also ensure implicit imitation can be applied widely. However, the degree to which rewards are the same and actions are homogeneous can have an impact on the utility (i.e., the acceleration of learning offered by) implicit imitation. We turn our attention to predicting the performance of implicit imitation as a function of certain domain characteristics.

## 6.1 Predicting Performance

In this section we examine two questions: first, given that implicit imitation is applicable, when can implicit imitation bias an agent to a suboptimal solution; and second, how will the performance of implicit imitation vary with structural characteristics of the domains one might want to apply it to? We show how analysis of the internal structure of state space can be used to motivate a metric that (roughly) predicts implicit imitation performance. We conclude with an analysis of how the problem space can be understood in terms of distinct regions playing different roles within an imitation context.

In the implicit imitation model, we use observations of other agents to improve the observer's knowledge about its environment and then rely on a sensible exploration policy to exploit this additional knowledge. A clear understanding of how knowledge of the environment affects exploration is therefore central to understanding how implicit imitation will perform in a domain.

Within the implicit imitation framework, agents know their reward functions, so knowledge of the environment consists solely of knowledge about the agent's action models. In general, these models can take any form. For simplicity, we have restricted ourselves to models that can be decomposed into local models for each possible combination of a system state and agent action.

The local models for state-action pairs allow the prediction of a $j$-step successor state distribution given any initial state and sequence of actions or local policy. The quality of the $j$-step state predictions will be a function of *every* action model encountered between the initial state and the states at time $j - 1$. Unfortunately, the quality of the $j$-step estimate can be drastically altered by the quality of even a single intermediate state-action model. This suggests that connected regions of state space, the states of which all have fairly accurate models, will allow reasonably accurate future state predictions.

Since the estimated value of a state $s$ is based on both the immediate reward and the reward expected to be received in subsequent states, the quality of this value estimate will also depend on the quality of the action models in those states connected to $s$. Now, since greedy exploration methods bias their exploration according to the estimated value of actions, the exploratory choices of an agent at state $s$ will also be dependent on the connectivity of reliable action models at those states reachable from $s$. Our analysis of implicit imitation performance with respect to domain characteristics is therefore organized around the idea of state space connectivity and the regions such connectivity defines.

### 6.1.1 THE IMITATION REGIONS FRAMEWORK

Since connected regions play an important role in implicit imitation, we introduce a classification of different regions within the state space shown graphically in Figure 21. In what follows, we describe of how these regions affect imitation performance in our model.

We first observe that many tasks can be carried out by an agent in a small subset of states within the state space defined for the problem. More precisely, in many MDPs, the optimal policy will ensure that an agent remains in a small subspace of state space. This leads us to the definition of our first regional distinction: relevant vs. irrelevant regions. The *relevant* region is the set of states with non-zero probability of occupancy under the optimal policy.[18] An $\varepsilon$-relevant region is a natural generalization in which the optimal policy keeps the system within the region a fraction $1 - \varepsilon$ of the time.

Within the relevant region, we distinguish three additional subregions. The *explored* region contains those states where the observer has formulated reliable action models on the basis of its own experience. The *augmented* region contains those states where the observer lacks reliable action models but has improved value estimates due to mentor observations.

---

18. One often assumes that the system starts in one of a small set of states. If the Markov chain induced by the optimal policy then is not ergodic, then the irrelevant region will be nonempty. Otherwise it will be empty.

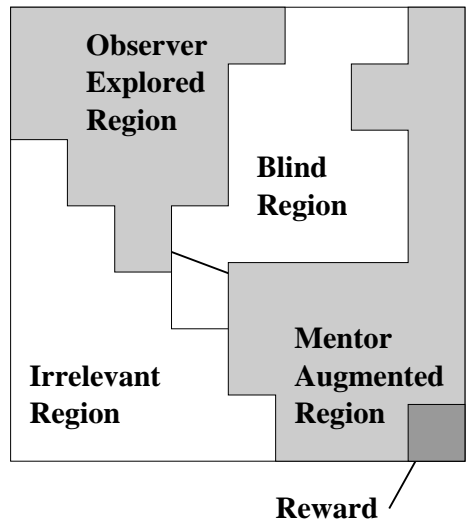

Figure 21: Classification of regions of state space

Note that both the explored and augmented regions are created as the result of observations made by the learner (of either its own transitions or those of a mentor). These regions will therefore have significant "connected components;" that is, contiguous regions of state space where reliable action or mentor models are available. Finally, the *blind* region designates those states where the observer has neither (significant) personal experience nor the benefit of mentor observations. Any information about states within the blind region will come (largely) from the agent's prior beliefs.[19]

We can now ask how these regions interact with an imitation agent. First we consider the impact of relevance. Implicit imitation makes the assumption that more accurate dynamics models allow an observer to make better decisions which will, in turn, result in higher returns sooner in the learning process. However, not all model information is equally helpful: the imitator needs only enough information about the irrelevant region to be able to avoid it. Since action choices are influenced by the relative value of actions, the irrelevant region will be avoided when it looks worse than the relevant region. Given diffuse priors on action models, none of the actions open to an agent will initially appear particularly attractive. However, a mentor that provides observations within the relevant region can quickly make the relevant region look much more promising as a method of achieving higher returns and therefore constrain exploration significantly. Therefore, considering problems just from the point of view of relevance, a problem with a small relevant region relative to the entire space combined with a mentor that operates within the relevant region will result in maximum advantage for an imitation agent over a non-imitating agent.

In the explored region, the observer has sufficiently accurate models to compute a good policy with respect to rewards within the explored region. Additional observations on the states within the explored region provided by the mentor can still improve performance somewhat if significant evidence is required to accurately discriminate between the expected value of two actions. Hence, mentor observations in the explored region can help, but will not result in dramatic speedups in convergence.

---

19. Our partitioning of states into explored, blind and augmented regions bears some resemblance to Kearns and Singh's (1998) partitioning of state space into known and unknown regions. Unlike Kearns and Singh, however, we use the partitions only for analysis. The implicit imitation algorithm does not explicitly maintain these partitions or use them in any way to compute its policy.

Now, we consider the augmented region in which the observer's Q-values have been augmented with observations of a mentor. In experiments in previous sections, we have seen that an observer entering an augmented region can experience significant speedups in convergence due to the information inherent in the augmented value function about the location of rewards in the region. Characteristics of the augmented zone, however, can affect the degree to which augmentation improves convergence speed.

Since the observer receives observations of only the mentor's state, and not its actions, the observer has improved value estimates for states in the augmented region, but no policy. The observer must therefore infer which actions should be taken to duplicate the mentor's behavior. Where the observer has prior beliefs about the effects of its actions, it may be able to perform immediate inference about the mentor's actual choice of action (perhaps using KL-divergence or maximum likelihood). Where the observer's prior model is uninformative, the observer will have to explore the local action space. In exploring a local action space, however, the agent must take an action and this action will have an effect. Since there is no guarantee that the agent took the action that duplicates the mentor's action, it may end up somewhere different than the mentor. If the action causes the observer to fall outside of the augmented region, the observer will lose the guidance that the augmented value function provides and fall back to the performance level of a non-imitating agent.

An important consideration, then, is the probability that the observer will remain in augmented regions and continue to receive guidance. One quality of the augmented region that affects the observer's probability of staying within its boundaries is its relative coverage of the state space. The policy of the mentor may be sparse or complete. In a relatively deterministic domain with defined begin and end states, a sparse policy covering few states may be adequate. In a highly stochastic domain with many start and end states, an agent may need a complete policy (i.e., covering every state). Implicit imitation will provide more guidance to the agent in domains that are more stochastic and require more complete policies, since the policy will cover a larger part of the state space.

As important as the completeness of a policy is in predicting its guidance, we must also take into account the probability of transitions into and out of the augmented region. Where the actions in a domain are largely invertible (directly, or effectively so), the agent has a chance of re-entering the augmented region. Where ergodicity is lacking, however, the agent may have to wait until the process undergoes some form of "reset" before it has the opportunity to gather additional evidence regarding the identity of the mentor's actions in the augmented region. The reset places the agent back into the explored region, from which it can make its way to the frontier where it last explored. The lack of ergodicity would reduce the agent's ability to make progress towards high-value regions before resets, but the agent is still guided on each attempt by the augmented region. Effectively, the agent will concentrate its exploration on the boundary between the explored region and the mentor augmented region.

The utility of mentor observations will depend on the probability of the augmented and explored regions overlapping in the course of the agent's exploration. In the explored regions, accurate action models allow the agent to move as quickly as possible to high value regions. In augmented regions, augmented Q-values inform agents about which states lead to highly-valued outcomes. When an augmented region abuts an explored region, the improved value estimates from the augmented region are rapidly communicated across the explored region by accurate action models. The observer can use the resultant improved value estimates in the explored region, together with the accurate action models in the explored region, to rapidly move towards the most promising states on the frontier of the explored region. From these states, the observer can explore outward and thereby eventually expand the explored region to encompass the augmented region.

In the case where the explored region and augmented region do not overlap, we have a blind region. Since the observer has no information beyond its priors for the blind region, the observer is reduced to random exploration. In a non-imitation context, any states that are not explored are blind. However, in an imitation context, the blind area is reduced in effective size by the augmented area. Hence, implicit imitation effectively shrinks the size of the search space of the problem even when there is no overlap between explored and augmented spaces.

The most challenging case for implicit imitation transfer occurs when the region augmented by mentor observations fails to connect to both the observer explored region and the regions with significant reward values. In this case, the augmented region will initially provide no guidance. Once the observer has independently located rewarding states, the augmented regions can be used to highlight "shortcuts". These shortcuts represent improvements on the agent's policy. In domains where a feasible solution is easy to find, but optimal solutions are difficult, implicit imitation can be used to convert a feasible solution to an increasingly optimal solution.

### 6.1.2 CROSS REGIONAL TEXTURES

We have seen how distinctive regions can be used to provide a certain level of insight into how imitation will perform in various domains. We can also analyze imitation performance in terms of properties that cut across the state space. In our analysis of how model information impacts imitation performance, we saw that regions connected by accurate action models allowed an observer to use mentor observations to learn about the most promising direction for exploration. We see, then, that any set of mentor observations will be more useful if it is concentrated on a connected region and less useful if dispersed about the state space in unconnected components. We are fortunate in completely observable environments that observations of mentors tend to capture continuous trajectories, thereby providing continuous regions of augmented states. In partially observable environments, occlusion and noise could lessen the value of mentor observations in the absence of a model to predict the mentor's state.

The effects of heterogeneity, whether due to differences in action capabilities in the mentor and observer or due to differences in the environment of the two agents, can also be understood in terms of the connectivity of action models. Value can propagate along chains of action models until we hit a state in which the mentor and observer have different action capabilities. At this state, it may not be possible to achieve the mentor's value and therefore, value propagation is blocked. Again, the sequential decision making aspect of reinforcement learning leads to the conclusion that many scattered differences between mentor and observer will create discontinuity throughout the problem space, whereas a contiguous region of differences between mentor and observer will cause discontinuity in a region, but leave other large regions fully connected. Hence, the *distribution pattern* of differences between mentor and observer capabilities is as important as the prevalence of difference. We will explore this pattern in the next section.

## 6.2 The Fracture Metric

We now try to characterize connectivity in the form of a metric. Since differences in reward structure, environment dynamics and action models that affect connectivity all would manifest themselves as differences in policies between mentor and observer, we designed a metric based on differences in the agents' optimal policies. We call this metric *fracture*. Essentially, it computes the average minimum distance from a state in which a mentor and observer disagree on a policy to a state in which mentor and observer agree on the policy. This measure roughly captures the difficulty the observer faces in profitably exploiting mentor observations to reduce its exploration demands.

More formally, let $\pi_m$ be the mentor's optimal policy and $\pi_o$ be the observer's. Let $\mathcal{S}$ be the state space and $\mathcal{S}_{\pi_m \neq \pi_o}$ be the set of *disputed* states where the mentor and observer have different optimal actions. A set of neighboring disputed states constitutes a disputed region. The set $\mathcal{S} - \mathcal{S}_{\pi_m \neq \pi_o}$ will be called the *undisputed states*. Let $M$ be a distance metric on the space $\mathcal{S}$. This metric corresponds to the number of transitions along the "minimal length" path between states (i.e., the shortest path using nonzero probability observer transitions).[20] In a standard grid world, it will correspond to the Manhattan distance. We define the *fracture* $\Phi(\mathcal{S})$ of state space $\mathcal{S}$ to be the average minimal distance between a disputed state and the closest undisputed state:

$$\Phi(\mathcal{S}) = \frac{1}{|\mathcal{S}_{\pi_m \neq \pi_o}|} \sum_{s \in \mathcal{S}_{\pi_m \neq \pi_o}} \min_{t \in \mathcal{S} - \mathcal{S}_{\pi_m \neq \pi_o}} M(s,t). \tag{13}$$

Other things being equal, a lower fracture value will tend to increase the propagation of value information across the state space, potentially resulting in less exploration being required. To test our metric, we applied it to a number of scenarios with varying fracture coefficients. It is difficult to construct scenarios which vary in their fracture coefficient yet have the same expected value. The scenarios in Figure 22 have been constructed so that the length of all possible paths from the start state $s$ to the goal state $x$ are the same in each scenario. In each scenario, however, there is an upper path and a lower path. The mentor is trained in a scenario that penalizes the lower path and so the mentor learns to take the upper path. The imitator is trained in a scenario in which the upper path is penalized and should therefore take the lower path. We equalized the difficulty of these problems as follows: using a generic $\varepsilon$-greedy learning agent with a fixed exploration schedule (i.e., a fixed initial rate and decay) in one scenario, we tuned the magnitude of penalties and their exact placement along loops in their other scenarios so that a learner using the same exploration policy would converge to the optimal policy in roughly the same number of steps in each.

---

20. The expected distance would give a more accurate estimate of fracture, but is more difficult to calculate.

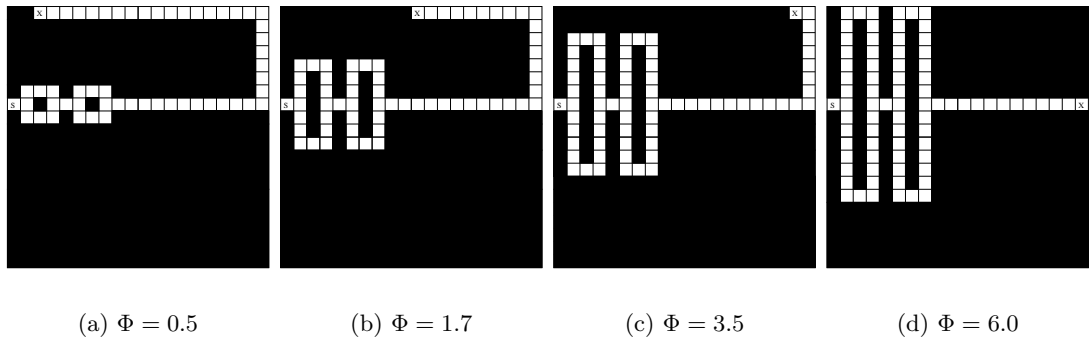

|     |     |     |     |     |
| --- | --- | --- | --- | --- |
| (a) $\Phi = 0.5$ | (b) $\Phi = 1.7$ | (c) $\Phi = 3.5$ | (d) $\Phi = 6.0$ |

Figure 22: Fracture metric scenarios

| $\Phi$ | Observer Initial Exploration Rate $\delta_I$ | | | | | | | |
| --- | --- | --- | --- | --- | --- | --- | --- | --- |
|     | $5 \times 10^{-2}$ | $1 \times 10^{-2}$ | $5 \times 10^{-3}$ | $1 \times 10^{-3}$ | $5 \times 10^{-4}$ | $1 \times 10^{-4}$ | $5 \times 10^{-5}$ | $1 \times 10^{-5}$ |
| 0.5 | 60% | 70% | 90% | | | | | |
| 1.7 | | | 0% | 80% | 90% | 90 % | | |
| 3.5 | | | | | 30% | 100 % | | |
| 6.0 | | | | | 30 % | 70 % | 100 % | 100 % |

Figure 23: Percentage of runs (of ten) converging to optimal policy given fracture $\Phi$ and initial exploration rate $\delta_I$

In Figure 22(a), the mentor takes the top of each loop and in an optimal run, the imitator would take the bottom of each loop. Since the loops are short and the length of the common path is long, the average fracture is low. When we compare this to Figure 22(d), we see that the loops are very long—the majority of states in the scenario are on loops. Each of these states on a loop has a distance to the nearest state where the observer and mentor policies agree, namely, a state not on the loop. This scenario therefore has a high average fracture coefficient.

Since the loops in the various scenarios differ in length, penalties inserted in the loops vary with respect to their distance from the goal state and therefore affect the total discounted expected reward in different ways. The penalties may also cause the agent to become stuck in a local minimum in order to avoid the penalties if the exploration rate is too low. In this set of experiments, we therefore compare observer agents on the basis of how likely they are to converge to the optimal solution given the mentor example.

Figure 23 presents the percentage of runs (out of ten) in which the imitator converged to the optimal solution (i.e., taking only the lower loops) as a function of exploration rate and scenario fracture.[21] We can see a distinct diagonal trend in the table illustrating that increasing fracture requires the imitator to increase its levels of exploration in order to find the optimal policy. This suggests that *fracture* reflects a feature of RL domains that is may be important in predicting the efficacy of implicit imitation.

---

21. For reasons of computational expediency, only the entries near the diagonal have been computed. Sampling of other entries confirms the trend.

## 6.3 Suboptimality and Bias

Implicit imitation is fundamentally about biasing the exploration of the observer. As such, it is worthwhile to ask when this has a positive effect on observer performance. The short answer is that a mentor following an optimal policy for an observer will cause an observer to explore in the neighborhood of the optimal policy and this will generally bias the observer towards finding the optimal policy.

A more detailed answer requires looking explicitly at exploration in reinforcement learning. In theory, an $\varepsilon$-greedy exploration policy with a suitable rate of decay will cause implicit imitators to eventually converge to the same optimal solution as their unassisted counterparts. However, in practice, the exploration rate is typically decayed more quickly in order to improve early exploitation of mentor input. Given practical, but theoretically unsound exploration rates, an observer may settle for a mentor strategy that is feasible, but non-optimal. We can easily imagine examples: consider a situation in which an agent is observing a mentor following some policy. Early in the learning process, the value of the policy followed by the mentor may look better than the estimated value of the alternative policies available to the observer. It could be the case that the mentor's policy actually is the optimal policy. On the other hand, it may be the case that one of the alternative policies, with which the observer has neither personal experience, nor observations from a mentor, is actually superior. Given the lack of information, an aggressive exploitation policy might lead the observer to falsely conclude that the mentor's policy is optimal. While implicit imitation can bias the agent to a suboptimal policy, we have no reason to expect that an agent learning in a domain sufficiently challenging to warrant the use of imitation would have discovered a better alternative. We emphasize that even if the mentor's policy is suboptimal, it still provides a feasible solution which will be preferable to no solution for many practical problems.

In this regard, we see that the classic exploration/exploitation tradeoff has an additional interpretation in the implicit imitation setting. A component of the exploration rate will correspond to the observer's belief about the sufficiency of the mentor's policy. In this paradigm, then, it seems somewhat misleading to think in terms of a decision about whether to "follow" a specific mentor or not. It is more a question of how much exploration to perform in addition to that required to reconstruct the mentor's policy.

## 6.4 Specific Applications

We see applications for implicit imitation in a variety of contexts. The emerging electronic commerce and information infrastructure is driving the development of vast networks of multi-agent systems. In networks used for competitive purposes such as trade, implicit imitation can be used by an RL agent to learn about buying strategies or information filtering policies of other agents in order to improve its own behavior.

In control, implicit imitation could be used to transfer knowledge from an existing learned controller which has already adapted to its clients to a new learning controller with a completely different architecture. Many modern products such as elevator controllers (Crites & Barto, 1998), cell traffic routers (Singh & Bertsekas, 1997) and automotive fuel injection systems use adaptive controllers to optimize the performance of a system for specific user profiles. When upgrading the technology of the underlying system, it is quite possible that sensors, actuators and the internal representation of the new system will be incompatible with the old system. Implicit imitation provides a method of transferring valuable user information between systems without any explicit communication.

A traditional application for imitation-like technologies lies in the area of bootstrapping intelligent artifacts using traces of human behavior. Research within the behavioral cloning paradigm has investigated transfer in applications such as piloting aircraft (Sammut et al., 1992) and controlling loading cranes (Šuc & Bratko, 1997). Other researchers have investigated the use of imitation to simplify the programming of robots (Kuniyoshi, Inaba, & Inoue, 1994). The ability of imitation to transfer complex, nonlinear and dynamic behaviors from existing human agents makes it particularly attractive for control problems.

## 7. Extensions

The model of implicit imitation presented above makes certain restrictive assumptions regarding the structure of the decision problem being solved (e.g., full observability, knowledge of reward function, discrete state and action space). While these simplifying assumptions aided the detailed development of the model, we believe the basic intuitions and much of the technical development can be extended to richer problem classes. We suggest several possible extensions in this section, each of which provides a very interesting avenue for future research.

### 7.1 Unknown Reward Functions

Our current paradigm assumes that the observer knows its own reward function. This assumption is consistent with the view of RL as a form of automatic programming. We can, however, relax this constraint assuming some ability to generalize observed rewards. Suppose that the expected reward can be expressed in terms of a probability distribution over features of the observer's state, $\Pr(r|f(s_o))$. In model-based RL, this distribution can be learned by the agent through its own experience. If the same features can be applied to the mentor's state $s_m$, then the observer can use what it has learned about the reward distribution to estimate expected reward for mentor states as well. This extends the paradigm to domains in which rewards are unknown, but preserves the ability of the observer to evaluate mentor experiences on its "own terms."

Imitation techniques designed around the assumption that the observer and the mentor share identical rewards, such as Utgoff's (1991), would of course work in the absence of a reward function. The notion of inverse reinforcement learning (Ng & Russell, 2000) could be adapted to this case as well. A challenge for future research would be to explore a synthesis between implicit imitation and reward inversion approaches to handle an observer's prior beliefs about some intermediate level of correlation between the reward function of observer and mentor.

## 7.2 Interaction of agents

While we cast the general imitation model in the framework of stochastic games, the restriction of the model presented thus far to noninteracting games essentially means that the standard issues associated with multiagent interaction do not arise. There are, of course, many tasks that require interactions between agents; in such cases, implicit imitation offers the potential to accelerate learning. A general solution requires the integration of imitation into more general models for multiagent RL based on stochastic or Markov games (Littman, 1994; Hu & Wellman, 1998; Bowling & Veloso, 2001). This would no doubt be a rather challenging, yet rewarding endeavor.

To take a simple example, in simple coordination problems (e.g., two mobile agents trying to avoid each other while carrying out related tasks) we might imagine an imitator learning from a mentor by reversing the roles of their roles when considering how the observed state transition is influenced by their joint action. In this and more general settings, learning typically requires great care, since agents learning in a nonstationary environment may not converge (say, to equilibrium). Again, imitation techniques offer certain advantages: for instance, mentor expertise can suggest means of coordinating with other agents (e.g., by providing a focal point for equilibrium selection, or by making clear a specific convention such as always "passing to the right" to avoid collision).

Other challenges and opportunities present themselves when imitation is used in multiagent settings. For example, in competitive or educational domains, agents not only have to choose actions that maximize information from exploration and returns from exploitation; they must also reason about how their actions communicate information to other agents. In a competitive setting, one agent may wish to disguise its intentions, while in the context of teaching, a mentor may wish to choose actions whose purpose is abundantly clear. These considerations must become part of any action selection process.

## 7.3 Partially Observable Domains

The extension of this model to partially observable domains is critical, since it is unrealistic in many settings to suppose that a learner can constantly monitor the activities of a mentor. The central idea of implicit imitation is to extract model information from observations of the mentor, rather than duplicating mentor behavior. This means that the mentor's internal belief state and policy are not (directly) relevant to the learner. We take a somewhat behaviorist stance and concern ourselves only with what the mentor's observed behaviors tell us about the possibilities inherent in the environment. The observer does have to keep a belief state about the mentor's current state, but this can be done using the same estimated world model the observer uses to update its own belief state.

Preliminary investigation of such a model suggests that dealing with partial observability is viable. We have derived update rules for augmented partially observable updates. These updates are based on a Bayesian formulation of implicit imitation which is, in turn, based on Bayesian RL (Dearden et al., 1999). In fully observable contexts, we have seen that more effective exploration using mentor observations is possible in fully observable domains when this Bayesian model of imitation is used (Price & Boutilier, 2003). The extension of this model to cases where the mentor's state is partially observable is reasonably straightforward. We anticipate that updates performed using a belief state about the mentor's state and action will help to alleviate fracture that could be caused by incomplete observation of behavior.

More interesting is dealing with an additional factor in the usual exploration-exploitation tradeoff: determining whether it is worthwhile to take actions that render the mentor "more visible" (e.g., ensuring the mentor remains in view so that this source of information remains available while learning).

## 7.4 Continuous and Model-Free Learning

In many realistic domains, continuous attributes and large state and action spaces prohibit the use of explicit table-based representations. Reinforcement learning in these domains is typically modified to make use of function approximators to estimate the Q-function at points where no direct evidence has been received. Two important approaches are parameter-based models (e.g., neural networks) (Bertsekas & Tsitsiklis, 1996) and the memory-based approaches (Atkeson, Moore, & Schaal, 1997). In both these approaches, model-free learning is generally employed. That is, the agent keeps a value function but uses the environment as an implicit model to perform backups using the sampling distribution provided by environment observations.

One straightforward approach to casting implicit imitation in a continuous setting would employ a model-free learning paradigm (Watkins & Dayan, 1992). First, recall the augmented Bellman backup function used in implicit imitation:

$$V(s) = R_o(s) + \gamma \max \left\{ \max_{a \in A_o} \left\{ \sum_{t \in S} \Pr_o(s, a, t) V(t) \right\}, \sum_{t \in S} \Pr_m(s, t) V(t) \right\} \quad (14)$$

When we examine the augmented backup equation, we see that it can be converted to a model-free form in much the same way as the ordinary Bellman backup. We use a standard Q-function with observer actions, but we will add one additional action which corresponds to the action $a_m$ taken by the mentor.[22] Now imagine that the observer was in state $s_o$, took action $a_o$ and ended up in state $s'_o$. At the same time, the mentor made the transition from state $s_m$ to $s'_m$. We can then write:

$$Q(s_o, a_o) = (1-\alpha)Q(s_o, a_o) + \alpha(R_o(s_o, a_o) + \gamma \max \left\{ \max_{a' \in A_o} \left\{ Q(s'_o, a') \right\}, Q(s'_o, a_m) \right\} \quad (15)$$

$$Q(s_m, a_m) = (1-\alpha)Q(s_m, a_m) + \alpha(R_o(s_m, a_m) + \gamma \max \left\{ \max_{a' \in A_o} \left\{ Q(s'_m, a') \right\}, Q(s'_m, a_m) \right\}$$

As discussed earlier, the relative quality of mentor and observer estimates of the Q-function at specific states may vary. Again, in order to avoid having inaccurate prior beliefs about the mentor's action models bias exploration, we need to employ a confidence measure to decide when to apply these augmented equations. We feel the most natural setting for these kind of tests is in the memory-based approaches to function approximation. Memory-based approaches, such as locally-weighted regression (Atkeson et al., 1997), not only provide estimates for functions at points previously unvisited, they also maintain the evidence set used to generate these estimates. We note that the implicit bias of memory-based approaches assumes smoothness between points unless additional data proves otherwise. On the basis of this bias, we propose to compare the average squared distance of the query from the exemplars used in the estimate of the mentor's Q-value to the average squared distance from the query to the exemplars used in the observer-based estimate to heuristically decide which agent has the more reliable Q-value.

---

22. This doesn't imply the observer knows which of *its* actions corresponds to $a_m$.

The approach suggested here does not benefit from prioritized sweeping. Prioritized-sweeping, has however, been adapted to continuous settings (Forbes & Andre, 2000). We feel a reasonably efficient technique could be made to work.

## 8. Related Work

Research into imitation spans a broad range of dimensions, from ethological studies, to abstract algebraic formulations, to industrial control algorithms. As these fields have cross-fertilized and informed each other, we have come to stronger conceptual definitions and a better understanding of the limits and capabilities of imitation. Many computational models have been proposed to exploit specialized niches in a variety of control paradigms, and imitation techniques have been applied to a variety of real-world control problems.

The conceptual foundations of imitation have been clarified by work on natural imitation. From work on apes (Russon & Galdikas, 1993), octopi (Fiorito & Scotto, 1992), and other animals, we know that socially facilitated learning is widespread throughout the animal kingdom. A number of researchers have pointed out, however, that social facilitation can take many forms (Conte, 2000; Noble & Todd, 1999). For instance, a mentor's attention to an object can draw an observer's attention to it and thereby lead the observer to manipulate the object independently of the model provided by the mentor. "True imitation" is therefore typically defined in a more restrictive fashion. Visalberghi and Fragazy (1990) cite Mitchell's definition:

1. something C (the copy of the behavior) is produced by an organism

2. where C is similar to something else M (the Model behavior)

3. observation of M is necessary for the production of C (above baseline levels of C occurring spontaneously)

4. C is designed to be similar to M

5. the behavior C must be a novel behavior not already organized in that precise way in the organism's repertoire.

This definition perhaps presupposes a cognitive stance towards imitation in which an agent explicitly reasons about the behaviors of other agents and how these behaviors relate to its own action capabilities and goals.

Imitation can be further analyzed in terms of the type of correspondence demonstrated by the mentor's behavior and the observer's acquired behavior (Nehaniv & Dautenhahn, 1998; Byrne & Russon, 1998). Correspondence types are distinguished by level. At the action level, there is a correspondence between actions. At the program level, the actions may be completely different but correspondence may be found between subgoals. At the effect level, the agent plans a set of actions that achieve the same effect as the demonstrated behavior but there is no direct correspondence between subcomponents of the observer's actions and the mentor's actions. The term *abstract imitation* has been proposed in the case where agents imitate behaviors which come from imitating the mental state of other agents (Demiris & Hayes, 1997).

The study of specific computational models of imitation has yielded insights into the nature of the observer-mentor relationship and how it affects the acquisition of behaviors by observers. For instance, in the related field of behavioral cloning, it has been observed that mentors that implement conservative policies generally yield more reliable clones (Urbancic & Bratko, 1994). Highly-trained mentors following an optimal policy with small coverage of the state space yield less reliable clones than those that make more mistakes (Sammut et al., 1992). For partially observable problems, learning from perfect oracles can be disastrous, as they may choose policies based on perceptions not available to the observer. The observer is therefore incorrectly biased away from less risky policies that do not require the additional perceptual capabilities (Scheffer, Greiner, & Darken, 1997). Finally, it has been observed that successful clones would often outperform the original mentor due to the "cleanup effect" (Sammut et al., 1992).

One of the original goals of behavioral cloning (Michie, 1993) was to extract knowledge from humans to speed up the design of controllers. For the extracted knowledge to be useful, it has been argued that rule-based systems offer the best chance of intelligibility (van Lent & Laird, 1999). It has become clear, however, that symbolic representations are not a complete answer. Representational capacity is also an issue. Humans often organize control tasks by time, which is typically lacking in state and perception-based approaches to control. Humans also naturally break tasks down into independent components and subgoals (Urbancic & Bratko, 1994). Studies have also demonstrated that humans will give verbal descriptions of their control policies which do not match their actual actions (Urbancic & Bratko, 1994). The potential for saving time in acquisition has been borne out by one study which explicitly compared the time to extract rules with the time required to program a controller (van Lent & Laird, 1999).

In addition to what has traditionally been considered imitation, an agent may also face the problem of "learning to imitate" or finding a correspondence between the actions and states of the observer and mentor (Nehaniv & Dautenhahn, 1998). A fully credible approach to learning by observation in the absence of communication protocols will have to deal with this issue.

The theoretical developments in imitation research have been accompanied by a number of practical implementations. These implementations take advantage of properties of different control paradigms to demonstrate various aspects of imitation. Early behavioral cloning research took advantage of supervised learning techniques such as decision trees (Sammut et al., 1992). The decision tree was used to learn how a human operator mapped perceptions to actions. Perceptions were encoded as discrete values. A time delay was inserted in order to synchronize perceptions with the actions they trigger. Learning apprentice systems (Mitchell et al., 1985) also attempted to extract useful knowledge by watching users, but the goal of apprentices is not to independently solve problems. Learning apprentices are closely related to programming by demonstration systems (Lieberman, 1993). Later efforts used more sophisticated techniques to extract actions from visual perceptions and abstract these actions for future use (Kuniyoshi et al., 1994). Work on associative and recurrent learning models has allowed work in the area to be extended to learning of temporal sequences (Billard & Hayes, 1999). Associative learning has been used together with innate following behaviors to acquire navigation expertise from other agents (Billard & Hayes, 1997).

A related but slightly different form of imitation has been studied in the multi-agent reinforcement learning community. An early precursor to imitation can be found in work on sharing of perceptions between agents (Tan, 1993). Closer to imitation is the idea of replaying the perceptions and actions of one agent for a second agent (Lin, 1991; Whitehead, 1991a). Here, the transfer is from one agent to another, in contrast to behavioral cloning's transfer from human to agent. The representation is also different. Reinforcement learning provides agents with the ability to reason about the effects of current actions on expected future utility so agents can integrate their own knowledge with knowledge extracted from other agents by comparing the relative utility of the actions suggested by each knowledge source. The "seeding approaches" are closely related. Trajectories recorded from human subjects are used to initialize a planner which subsequently optimizes the plan in order to account for differences between the human effector and the robotic effector (Atkeson & Schaal, 1997). This technique has been extended to handle the notion of subgoals within a task (Atkeson & Schaal, 1997). Subgoals are also addressed by others (Šuc & Bratko, 1997). Our own work is based on the idea of an agent extracting a model from a mentor and using this model information to place bounds on the value of actions using its *own* reward function. Agents can therefore learn from mentors with reward functions different than their own.

Another approach in this family is based on the assumption that the mentor is rational (i.e., follows an optimal policy), has the same reward function as the observer and chooses from the same set of actions. Given these assumptions, we can conclude that the action chosen by a mentor in a particular state must have higher value to the mentor than the alternatives open to the mentor (Utgoff & Clouse, 1991) and therefore higher value to the observer than any alternative. The system of Utgoff and Clouse therefore iteratively adjusts the values of the actions until this constraint is satisfied in its model. A related approach uses the methodology of linear-quadratic control (Šuc & Bratko, 1997). First a model of the system is constructed. Then the inverse control problem is solved to find a cost matrix that would result in the observed controller behavior given an environment model. Recent work on inverse reinforcement learning takes a related approach to reconstructing reward functions from observed behavior (Ng & Russell, 2000). It is similar to the inversion of the quadratic control approach, but is formulated for discrete domains.

Several researchers have picked up on the idea of common representations for perceptual functions and action planning. One approach to using the same representation for perception and control is based on the PID controller model. The PID controller represents the behavior. Its output is compared with observed behaviors in order to select the action which is closest to the observed behavior (Demiris & Hayes, 1999). Explicit motor action schema have also been investigated in the dual role of perceptual and motor representations (Matarić, Williamson, Demiris, & Mohan, 1998).

Imitation techniques have been applied in a diverse collection of applications. Classical control applications include control systems for robot arms (Kuniyoshi et al., 1994;

Friedrich, Munch, Dillmann, Bocionek, & Sassin, 1996), aeration plants (Scheffer et al., 1997), and container loading cranes (Šuc & Bratko, 1997; Urbancic & Bratko, 1994). Imitation learning has also been applied to acceleration of generic reinforcement learning (Lin, 1991; Whitehead, 1991a). Less traditional applications include transfer of musical style (Cañamero, Arcos, & de Mantaras, 1999) and the support of a social atmosphere (Billard, Hayes, & Dautenhahn, 1999; Breazeal, 1999; Scassellati, 1999). Imitation has also been investigated as a route to language acquisition and transmission (Billard et al., 1999; Oliphant, 1999).

## 9. Concluding Remarks

We have described a formal and principled approach to imitation called implicit imitation. For stochastic problems in which explicit forms of communication are not possible, the underlying model-based framework combined with model extraction provides an alternative to other imitation and learning-by-observation systems. Our new approach makes use of a model to compute the actions an imitator should take without requiring that the observer duplicate the mentor's actions exactly. We have shown implicit imitation to offer significant transfer capability on several test problems, where it proves to be robust in the face of noise, capable of integrating subskills from multiple mentors, and able to provide benefits that increase with the difficulty of the problem.

We have seen that feasibility testing extends implicit imitation in a principled manner to deal with the situations where the homogeneous action assumption is invalid. Adding bridging capabilities preserves and extends the mentor's guidance in the presence of infeasible actions, whether due to differences in action capabilities or local differences in state spaces. Our approach also relates to the idea of "following" in the sense that the imitator uses local search in its model to repair discontinuities in its augmented value function before acting in the world. In the process of applying imitation to various domains, we have learned more about its properties. In particular we have developed the *fracture* metric to characterize the effectiveness of a mentor for a given observer in a specific domain. We have also made considerable progress in extending imitation to new problem classes. The model we have developed is rather flexible and can be extended in several ways: for example, a Bayesian approach to imitation building on this work shows great potential (2003); and we have initial formulations of promising approaches to extending implicit imitation to multi-agent problems, partially observable domains and domains in which the reward function is not specified a priori.

A number of challenges remain in the field of imitation. Bakker and Kuniyoshi (1996) describe a number of these. Among the more intriguing problems unique to imitation are: the evaluation of the expected payoff for observing a mentor; inferring useful state and reward mappings between the domains of mentors and those of observers; and repairing or locally searching in order to fit observed behaviors to an observer's own capabilities and goals. We have also raised the possibility of agents attempting to reason about the information revealed by their actions in addition to whatever concrete value the actions have for the agent.

Model-based reinforcement has been applied to numerous problems. Since implicit imitation can be added to model-based reinforcement learning with relatively little effort, we expect that it can be applied to many of the same problems. Its basis in the simple but elegant theory of Markov decision processes makes it easy to describe and analyze. Though we have focused on some simple examples designed to illustrate the different mechanisms required for implicit imitation, we expect that variations on our approach will provide interesting directions for future research.

## Acknowledgments

Thanks to the anonymous referees for their suggestions and comments on earlier versions of this work and Michael Littman for editorial suggestions. Price was supported by NCE IRIS-III Project BAC. Boutilier was supported by NSERC Research Grant OGP0121843, and the NCE IRIS-III Project BAC. Some parts of this paper were presented in "Implicit Imitation in Reinforcement Learning," *Proceedings of the Sixteenth International Conference on Machine Learning (ICML-99)*, Bled, Slovenia, pp.325–334 (1999) and "Imitation and Reinforcement Learning in Agents with Heterogeneous Actions," *Proceedings Fourteenth Biennial Conference of the Canadian Society for Computational Studies of Intelligence (AI 2001)*, Ottawa, pp.111–120 (2001).